\documentclass[lettersize,journal]{IEEEtran}
\IEEEoverridecommandlockouts

\usepackage{amsmath,amsfonts}
\usepackage{algorithmic}
\usepackage{array}
\usepackage[caption=false,font=normalsize,labelfont=sf,textfont=sf]{subfig}
\usepackage{textcomp}
\usepackage{stfloats}
\usepackage{url}
\usepackage{verbatim}
\usepackage{graphicx}
\usepackage{cite}
\usepackage{bm}
\usepackage{xcolor}
\definecolor{linkblue}{RGB}{0, 70, 140}
\usepackage{booktabs}
\usepackage{multirow}
\usepackage{stackengine}
\usepackage{pifont}
\usepackage{amssymb}
\usepackage{arydshln}
\def\BibTeX{{\rm B\kern-.05em{\sc i\kern-.025em b}\kern-.08em
    T\kern-.1667em\lower.7ex\hbox{E}\kern-.125emX}}
\usepackage{balance}
\usepackage[
    colorlinks=true,
    allcolors=linkblue
]{hyperref}

\begin{document}
\title{Physics-Guided Robotic Radiation Source Localization along Arbitrary Measurement Paths in Unstructured Environments}
\author{Hojoon Son$^{1}$, \textit{Graduate Student Member, IEEE,} Kai Tan$^{1}$, and Fan Zhang$^{1,*}$, \textit{Senior Member, IEEE}%
\thanks{$^{1}$ George W. Woodruff School of Mechanical Engineering, Georgia Institute of Technology, Atlanta, GA, 30332-0405 USA (e-mail: {\tt\small hojoon.son@gatech.edu}, {\tt\small ktan79@gatech.edu}, {\tt\small fan@gatech.edu})}
\thanks{$^{*}$ Corresponding Author: Fan Zhang}
}

\markboth{Published in IEEE Transactions on Automation Science and Engineering.}{}

\IEEEpubid{%
  \begin{minipage}{\textwidth}
    \centering
    \footnotesize
    \copyright~2026 IEEE.  Personal use of this material is permitted.  Permission from IEEE must be obtained for all other uses, in any current or future media, including reprinting/republishing this material for advertising or promotional purposes, creating new collective works, for resale or redistribution to servers or lists, or reuse of any copyrighted component of this work in other works.
  \end{minipage}%
}

\maketitle

\IEEEpubidadjcol

\begin{abstract}
Using robots to estimate the location of the radiation source is an effective way to improve efficiency and safety. Existing methods focus on planning the robot's path to achieve precise estimation, typically approaching the source. However, approaching the source increases the risk of radiation damage to a robot. In addition, a path-planning algorithm designed solely for radiation source localization (RSL) limits the flexibility of missions that deploy robots into radioactive environments. This study presents an automation framework for robotic RSL that leverages a physics-informed machine learning (PIML) model to precisely estimate the source location, regardless of measurement paths, in unknown environments. Physics-inspired model tensors have been designed for PIML to handle attenuated gamma-ray flux signals from unknown obstacles, and multiple models are computed in parallel to improve the robustness and precision of the RSL. The proposed method is evaluated in high-fidelity simulation environments using Monte Carlo particle transport across diverse randomized domains, including spatial scales, radiation source types, obstacle materials and geometries, and robot trajectories. The method is also validated through physical experiments on configurations that are not included in the simulation-based evaluation. The continuous learning technique is applied in real-robot deployment to enhance the practical applicability of the online robotic RSL system. The proposed method advances robot radiation perception from pointwise flux detection to spatial intelligence.
\end{abstract}

\def\abstractname{Note to Practitioners}
\begin{abstract}
This paper presents a framework for localizing a radiation source using sparse measurements collected by a robot. The proposed method is an independent module, regardless of the robot’s target mission in radioactive environments; it does not actively plan a trajectory solely to localize a radiation source. In addition, the method handles complex signals affected by detection noise and obstacle-induced attenuation in unstructured environments. The proposed framework has been validated not only in simulation but also in physical setups. The robotic radiation detection system, which integrates the RadEye SPRD-ER radiation detector with the Unitree Go2 EDU quadruped robot, is used for physical experiments in laboratory environments. Demonstrating and validating this method in large-scale practical environments is a remaining challenge for industry-level deployment.
\end{abstract}


\section{Introduction}
\IEEEPARstart{R}{adioactive} environments are safety-critical domains where robotic automation, such as for radiation surveying, is primarily beneficial. Traditionally, radiation surveys have been conducted by human operators carrying detectors and collecting multiple flux levels. These can be replaced with mobile robots for efficiency and safety. Specifically, robotic automation for estimating the location of a radiation source is a feasible and straightforward solution.

Although using robots can be a safer option, several factors make robotic radiation source localization (RSL) highly complex and challenging. These challenges include Poisson detection noise, high complexity in the radiation field due to the superposition of multiple sources, and attenuation from shielding materials~\cite{okabe2024tetris,son2025physics}. Another challenge of robotic RSL is minimizing radiation exposure for a robot from an As Low As Reasonably Practicable (ALARP) perspective~\cite{abrahamsen2007use}. Previous studies have investigated active search methods that greedily select robotic actions that maximize information gain for RSL~\cite{huo2020autonomous,lazna2025localizing,zhang2025radiation,nguyen2024radiation,zhu2024dual}. However, adaptive planning with active search causes robots to gradually approach the radiation source, which is potentially unsafe and increases the chance of radiation damage to robots~\cite{son2025physics}.

Moreover, low mission flexibility is a critical bottleneck of practical robotic RSL. Estimating source locations is not the only robotic mission in nuclear environments. For example, recent research proposes robotic use cases for certain nuclear power plant (NPP) operations and maintenance (O\&M), and proposed tasks are not limited to radiation surveys, but also include operational monitoring and inspection~\cite{son2025use}. Therefore, a path-planning method designed solely for RSL prevents robots from performing general tasks in radioactive environments, which is highly inefficient in practice. 

In this context, an RSL method that is independent of the trajectory used for radiation flux acquisition in diverse missions beyond radiation survey will be advantageous. With this method, estimated source locations can serve as ``advanced perception information" derived from pure flux data. Additionally, this method helps mitigate the risk of radiation exposure across diverse missions in critical environments. As the method does not actively approach the radiation source, the low-risk robot path based on this ``advanced perception information" is intended to be planned by the user.

\IEEEpubidadjcol
\IEEEpubidadjcol

Physics-informed machine learning (PIML) is a promising approach for precise estimation of radiation source locations~\cite{son2025physics}. However, previous studies limited the target environment to air alone, thereby neglecting obstacle-induced attenuation. Also, the previous research constrained the source type and the robot's path to acquire radiation-flux data. Furthermore, the previous study did not perform generalized validations across diverse conditions and did not conduct physical experiments with actual sources, detectors, and robots. 

This paper presents the PIML-based robotic RSL method, which uses flux measurements acquired along arbitrary robot paths, ensuring its applicability to any mission in radioactive environments. It successfully handles complexity from various types of radiation sources and unknown obstacle-induced attenuation. The method is validated in cross-condition unstructured environments that include varying radiation source types, as well as multiple obstacle materials, shapes, and sizes. The main contributions and innovations are as follows.

\begin{itemize}
    \item A PIML-based robotic RSL framework, structured around physics-inspired tensors, offset tensors, and parallel inference, has been developed. It outperforms previous methods by explicitly addressing both inherent detection uncertainty and obstacle-induced attenuation in unknown environments, regardless of background radiation, radiation source type, obstacle properties, or spatial geometry.
    \item The innovation of the proposed RSL framework is its applicability to unplanned measurement trajectories. Unlike active path planning solely for finding a radiation source, the proposed method can serve as an independent RSL module for various robotic missions.
    \item The proposed approach was validated in unstructured environments of various settings, featuring heterogeneous obstacle materials and geometries, and a wide range of radiation source types. The validation encompasses both high-fidelity Monte Carlo simulations and real-world experiments, confirming its precision, robustness, and practical applicability.
\end{itemize}

Section~\ref{section:related_works} provides a review of previous related studies and highlights the advantages of this research. Section~\ref{section:methodology} explains the detailed methodology of the proposed robotic RSL framework. Section~\ref{section:sim_results} and \ref{section:experimental_validation} present the results of simulations and physical experiments, respectively. Section~\ref{section:discussion} discusses the proposed framework, including limitations and future directions. In Section~\ref{section:conclusion}, the conclusion is presented.

\section{Related Works}
\label{section:related_works}
In obstacle-free environments, the space is filled only with air, making the RSL problem simpler than in obstacle-populated environments. Traditional RSL approaches were based on gradient descent~\cite{baidoo2013gradient} with a small number of detectors, triangulation~\cite{grieme2015triangulation} using geometric and mathematical relationships among multiple measurement points, and maximum likelihood estimation (MLE) with unknown source signal statistics~\cite{baidoo2016maximum}. More recent methods for generic signal source search have leveraged information theory, such as infotaxis~\cite{park2022receding} and entrotaxis~\cite{hutchinson2018entrotaxis}. Huo~\textit{et al.} suggested the RSL method, applying a partially observable Markov decision process (POMDP) to select the robot behavior and estimating parameters with a Bayesian framework~\cite{huo2020autonomous}. One of the most recent RSL methods in obstacle-free environments is based on PIML. Son~\textit{et al.} proposed a multiple radiation source localization method by combining a data-driven source count classifier and a physics model-based optimizer~\cite{son2025physics}. However, Son's method restricts the measurement trajectory and does not account for environments with obstacles.

The robotic RSL will be more challenging in obstacle-populated environments because the measurement sequence along the robot path will be complicated when measurements are attenuated. Still, if the obstacle properties are specified, the attenuation coefficients can be determined, thereby simplifying the RSL problem compared with the case where the obstacle features are unknown. The particle filter (PF) is a powerful method for RSL in robot exploration~\cite{pinkam2020informative}. Kemp~\textit{et al.} proposed a PF-based approach to estimate multiple radiation sources in obstacle-rich environments~\cite{kemp2023real}. After attenuation is computed from the local terrain and obstacle data, dynamic particle allocation is employed to localize radiation sources. Lazna and Zalud also leveraged PF for RSL with active robot trajectory planning~\cite{lazna2025localizing}. In addition, PF was used to estimate radiation source locations through heterogeneous multi-robot collaboration~\cite{ling2025heterogeneous}. Nguyen~\textit{et al.} presented a localization method by estimating a high existence possibility area using an extended Kalman filter (EKF), and then finding a radiation source's location with the maximum likelihood expectation maximization (ML-EM) technique~\cite{nguyen2024radiation}. Nguyen's method integrates 3D object recognition to estimate the probability of radiation shielding for the detected object, and geometries and materials of detected obstacles are corrected by a prior known blueprint of the environment. Another approach is to use simultaneous localization and mapping (SLAM) to scan the 3D shapes of obstacles~\cite{bandstra2021improved}, or to use multi-kernel Gaussian process regression (MK-GPR) to map the radiation distribution~\cite{zhang2025radiation}. However, these methods are available only when the material properties and geometric settings of the obstacles are known.

The ultimate challenge in robotic RSL arises when both the obstacle properties and source parameters are entirely undefined. One solution is to use an advanced hardware device, a Compton camera, instead of a flux-intensity-based radiation detector~\cite{lee2018active,baca2021gamma,werner2024autonomous}. However, a Compton camera is too expensive for widespread deployment in robotic RSL~\cite{kagaya2015development}. Rather than relying on advanced hardware, some studies have proposed methodological frameworks to address this challenge. While Nguyen’s method~\cite{nguyen2024radiation} was primarily designed for known environments with blueprints, it can also be extended to unknown environments, assuming that 3D object detection can precisely recognize obstacle geometries and accurately predict material properties. Zhu~\textit{et al.} suggested a dual-stage search method by alternatively conducting the source tracking stage and relocation stage to rapidly search radioactive sources in unknown environments~\cite{zhu2024dual}. They divided the obstacle-populated space into obstacle-free convex polyhedra to filter out pseudo-sources in complex environments. However, previous studies did not account for the varied shapes and sizes of obstacles or the heterogeneous material types. Additionally, they did not present a systematic performance evaluation using a large amount of highly complex, multiscale data that accounted for robots' radiation exposure safety.

\begin{figure*}[t]
    \centering
    \includegraphics[width=.95\textwidth]{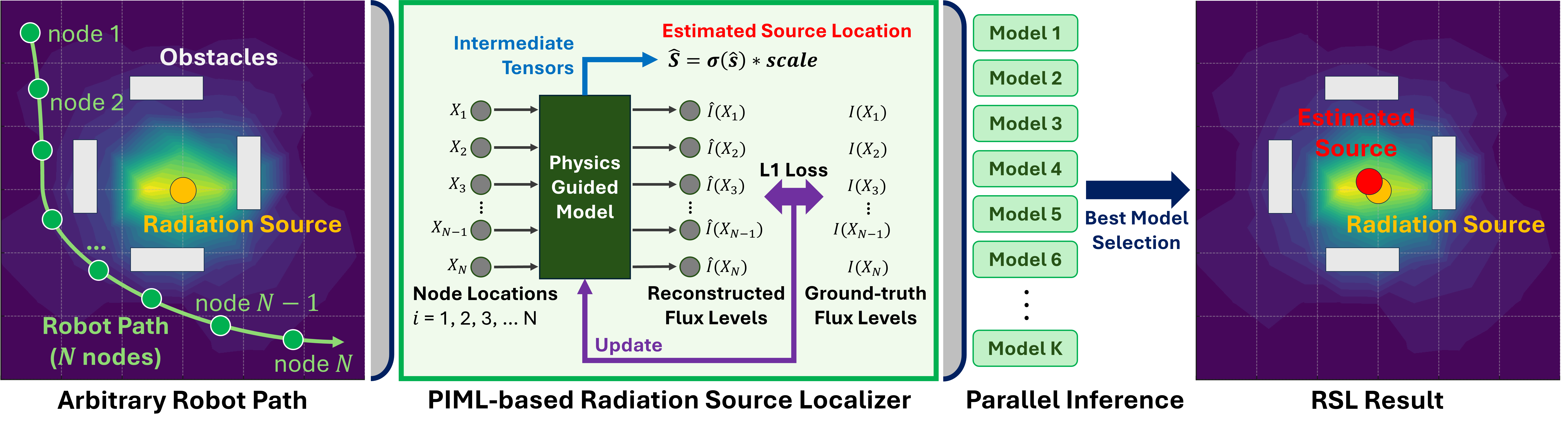}
    \caption{Proposed robotic RSL framework}
    \label{fig:overall}
\end{figure*}

\section{Methodology}
\label{section:methodology}

Figure~\ref{fig:overall} illustrates the proposed robotic RSL framework. In an unknown environment with unstructured obstacles, the robot moves along an arbitrary path while measuring gamma-ray flux. The PIML-based radiation source localizer is designed using physics-inspired trainable tensors. The model is trained to minimize the flux reconstruction error, and the estimated source location is extracted from the model's intermediate physics-inspired tensors. Multiple RSL processes are implemented in parallel. The final output is obtained from the best-performing model, which has the lowest loss, among the multiple inferences. Parallel inference improves the accuracy and robustness of the RSL.

\subsection{Physics Model}
This research focuses on gamma-ray detection. There are two reasons to use gamma particles as a key signal for robotic RSL: (i) Gamma radiation is a common byproduct of most radioactive sources. Therefore, detecting gamma radiation is a reasonable approach for localizing a wide range of radiation sources. (ii) Gamma radiation has high penetrating power, making it theoretically detectable at long distances from the source, which makes the robotic RSL feasible.

The fundamental physics model of gamma-ray distribution is a combination of the Beer-Lambert law and the inverse-square law~\cite{lambert1760photometria,knoll2010radiation}. According to the Beer-Lambert law, the amount of light absorption is proportional to the length and density of the medium it passes through. It can be mathematically expressed as an exponential function. Following the inverse-square law, the radiation flux decreases in inverse proportion to the square of the distance from the source. In the practical radiation detection with a detector, the inverse-square law can be expressed with the detector radius~\cite{knoll2010radiation,wright2021simulating}.

Equation~\eqref{eqn:gamma_physics} represents the gamma-ray flux model proposed by Wright~\textit{et al.}~\cite{wright2021simulating}, combining the Beer-Lambert term and detector radius-included inverse-square term. The equation closely follows the Monte Carlo N-Particle 6 (MCNP6) code~\cite{werner2018mcnp} and is applicable to robotic radiation detection in a Gazebo-based simulation environment. In this formulation, $I$ and $I_0$ represent the detected and incident intensities, respectively; $\eta$ and $r$ denote the sensitivity and radius of the detector; $d$ is the distance between the source and the detection point; and $\mu_j$ and $x_j$ are the attenuation coefficient and thickness of the $j$-th medium encountered along the gamma-ray path. Suppose that the line of sight (LoS) between the source and the detection point intersects $k$ distinct intersecting media. 
\begin{equation}
    I =  \eta \frac{I_{0}}{2} \left(1 - \frac{d}{\sqrt{r^2 + d^2}} \right) \prod_{j=1}^k \exp (-\mu_j x_j)
    \label{eqn:gamma_physics}
\end{equation}

The equation can be extended by applying a build-up factor to model the scattering of gamma radiation~\cite{sardari2009estimation,zuo2018comparative,das2022determination}. The build-up factor is determined by the gamma-ray energy and the properties of the shielding material. Equation~\eqref{eqn:gamma_physics_buildup} shows the final formula to model the gamma-ray flux regarding the inverse-square law, detector geometry, attenuation, and scattering, where $B_j$ is a build-up factor of the $j$-th medium.
\begin{equation}
    I =  \eta \frac{I_{0}}{2} \left(1 - \frac{d}{\sqrt{r^2 + d^2}} \right) \prod_{j=1}^k B_j \exp (-\mu_j x_j)
    \label{eqn:gamma_physics_buildup}
\end{equation}
\subsection{PIML Model Design}
\label{sec:piml_model_design}
The proposed PIML model employs a hybrid architecture, defined as a weighted sum of the obstacle-free and obstacle-aware models. The obstacle-free model and obstacle-aware model partially share parameters, and the weight $w$ of the hybrid model is a trainable parameter. All physics parameters are formulated as tensors to be incorporated into the hybrid PIML model.

\subsubsection{Obstacle-free Model}
With the number of flux measurements $N$ along the robot path, let the $i$-th measurement location $\mathbf{X_i} = (x_i, y_i) \in \mathbb{R}^2$. Also, let the ground-truth source location $\mathbf{S} = (S_x, S_y) \in \mathbb{R}^2$ and the estimated source location $\mathbf{\hat{S}} = (\hat{S_x}, \hat{S_y}) \in \mathbb{R}^2$. To constrain the estimated values within the unit interval $(0,1)$, the trainable parameter $\hat{s} = (\hat{s}_x, \hat{s}_y)$ is passed through the sigmoid function~\eqref{eqn:sigmoid}, and then rescaled by the environment's dimension to map the estimated source location back to the original coordinate system. Equation~\eqref{eqn:S_xy} represents this process.
\begin{equation}
    \sigma(x) = \frac{1}{1+e^{-x}}
    \label{eqn:sigmoid}
\end{equation}
\begin{equation}
    \hat{S_x} = \sigma(\hat{s}_x) \cdot \text{scale}_x, \quad \hat{S_y} = \sigma(\hat{s}_y) \cdot \text{scale}_y
    \label{eqn:S_xy}
\end{equation}
The estimated distance from the flux measurement location and the estimated source location is calculated using the Euclidean distance formula, as represented in equation~\eqref{eqn:distance}.
\begin{equation}
    \hat{d_i} = \sqrt{(x_i - \hat{S_x})^2 + (y_i - \hat{S_y})^2} - \operatorname{Avg}(O)
    \label{eqn:distance}
\end{equation}

One of the major findings in this paper is that RSL performance is significantly improved when the offset values, $O=\{o_1, o_2, o_3\}$, are incorporated into the estimated distance. Physically, the offset is necessary in the inverse-square formula to prevent an infinite value at the source location. Additionally, it makes model parameter estimation robust by intentionally tolerating small errors in distance estimation. Sigmoid-applied $\hat{s}$ passes through the multi-layer perceptron (MLP), and three offset values are retrieved, as shown in equation~\eqref{eqn:offset}. A mean of offsets is subtracted from the Euclidean distance between the estimated source and the measurement location, as represented in equation~\eqref{eqn:distance}. $\operatorname{Avg}(O)$ denotes the arithmetic mean of the offset value set $O$.
\begin{equation}
    O = \{o_1, o_2, o_3\} = \text{MLP}(\sigma(\hat{s_x}), \sigma(\hat{s_y}))
    \label{eqn:offset}
\end{equation}
Each offset value is re-added to $\hat{d}_i$ individually to get distinct offset-applied distances. Equation~\eqref{eqn:offset_dist} represents this process.
\begin{equation}
    \hat{D}_{i,j} = \hat{d_i} + o_j, \quad j \in \{1, 2, 3\}
    \label{eqn:offset_dist}
\end{equation}

Equation~\eqref{eqn:obstacle-free} shows the obstacle-free model based on the physics model described previously by equation~\eqref{eqn:gamma_physics_buildup}, where $R_{\text{free}}$ denotes a detector radius. Note that the build-up factor can be ignored, and the air attenuation coefficient can be fixed to a specific value in the obstacle-free environment. The air attenuation coefficient $\mu_{air}$ is fixed to $9.076\times10^{-5}\mathrm{cm}^{-1}$ from the reference~\cite{wright2021simulating}. The term $\eta \frac{I_{0}}{2}$ is combined into the squared term $P^2>0$ to ensure the value is positive. 
\begin{equation}
    \hat{\mathbf{I}}_{\text{free}}(\mathbf{X}_i) = P_{\text{free}}^2\left(1-\frac{\hat{D}_{i,1}}{\sqrt{{R}_{\text{free}}^2 + \hat{D}_{i,2}^2}} \right) \exp (-\mu_{\text{air}}\cdot\hat{D}_{i,3}) 
    \label{eqn:obstacle-free}
\end{equation}
\subsubsection{Obstacle-aware Model}
The obstacle-aware model considers attenuation and scattering across heterogeneous materials in the environment. In this case, build-up factors ($B_j$), attenuation coefficients ($\mu_j$), and thicknesses ($x_j$) should be estimated across $k$ intersected obstacles. However, the major assumption of this research is that RSL is conducted in unknown environments. Therefore, the model does not know the number, geometry, location, or materials of the obstacles; therefore, estimating $B$, $\mu$, and $x$ is not feasible. To resolve this challenge, the term ($\prod_{j=1}^k B_j \exp(-\mu_j x_j)$) is combined into one parameter, as it is represented in equation~\eqref{eqn:A_i}. Although ($\ln(B) - \mu x$) is not mathematically guaranteed to be negative, the physical flux reduction by attenuation and scattering justifies the combined term as a single negative parameter, $-A$. Estimating $A$ is physically equivalent to the joint estimation of $B, \mu$, and $x$ for multiple unknown obstacles.
\begin{align}
    \prod_{j=1}^k B_j \exp(-\mu_j x_j) &= \exp \left( \sum_{j=1}^k \ln(B_j) - \sum_{j=1}^k \mu_j x_j \right) \notag \\ 
    &= \exp \left( \sum_{j=1}^k \left( \ln(B_j) - \mu_j x_j \right) \right) \notag \\
    &\approx \exp(-A), \quad (A > 0)
    \label{eqn:A_i}
\end{align}

For each flux measurement, the value is affected by attenuation, which depends on the source location, detection point, and environmental geometry; some flux measurements are attenuated, whereas others are not. To estimate those, $\exp(-A)$ can be conditionally fitted. If the $i$-th measurement is attenuated, $A_i$ will be a positive number; otherwise, $A_i$ will be zero. Equation~\eqref{eqn:condition_A} shows the conditional square activation function to determine $A_i$ by fitting the trainable parameter $a_i$. This discrete function is optimized using the customized straight-through estimator (STE). 
\begin{equation}
    A_i = 
    \begin{cases} 
        {a_i}^2 & \text{if } a_i > 0 \\
        0 & \text{otherwise}
    \end{cases}
    \label{eqn:condition_A}
\end{equation}

Equation~\eqref{eqn:obstacle} shows the obstacle-aware model. It accounts for the masked attenuation parameter $A_i$. Distance parameters are shared with the obstacle-free model, and $P_{\text{obs}}$ and $R_{\text{obs}}$ are separated. 
\begin{equation}
    \hat{\mathbf{I}}_{\text{obs}}(\mathbf{X}_i) = P_{\text{obs}}^2\left(1-\frac{\hat{D}_{i,1}}{\sqrt{{R}_{\text{obs}}^2 + \hat{D}_{i,2}^2}} \right) \exp (-\mu_{\text{air}}\cdot\hat{D}_{i,3} - A_i) 
    \label{eqn:obstacle}
\end{equation}
\subsubsection{Hybrid Model}
In the ideal scenario, only using the obstacle-aware model is enough if masked attenuation parameters $A_i, (i=1, 2, ..., N)$ are estimated precisely. However, the obstacle-free model is more accurate and robust than the obstacle-aware model in environments with few obstacles, whereas the obstacle-aware model is better than the obstacle-free model in environments with dense obstacles. Therefore, the hybrid architecture is leveraged to balance each model's advantage.

Equation~\eqref{eqn:hybrid} represents the proposed hybrid PIML model by balancing the weight of the obstacle-free model and the obstacle-aware model. To ensure the weight value is between 0 and 1, apply the sigmoid activation function to the trainable weight parameter $w$.
\begin{align}
    \hat{\mathbf{I}}(\mathbf{X}_i) = \sigma(w) \cdot \hat{\mathbf{I}}_{\text{free}}(\mathbf{X}_i) + (1 - \sigma(w)) \cdot \hat{\mathbf{I}}_{\text{obs}}(\mathbf{X}_i) \label{eqn:hybrid} \\
    i = 1, 2, \dots, N \nonumber
\end{align}

\subsubsection{Optimization}
\label{section:optimization}
All trainable tensors at the epoch $t$ are represented as $\theta_t$. The values of $P$ and $R$ for obstacle-free and obstacle-aware models are randomly initialized to values between 0 and 1. The values of $a_i, (i = 1, 2, ... N)$ and $w$ are initialized as 0. The values of $\hat{s}_x$ and $\hat{s}_y$ are randomly initialized to values between -0.1 and 0.1, to ensure the sigmoid-applied values are between 0.475 and 0.525.  

The detected flux sequence is normalized by dividing each flux value by its maximum, as shown in equation~\eqref{eqn:max_scaling}. This scaling process prevents the PIML model from diverging across various unstructured environmental setups.
\begin{equation}
\mathbf{I'}=\frac{\mathbf{I}}{\text{max}(\mathbf{I})}
\label{eqn:max_scaling}
\end{equation}
$\theta_t$ is optimized using the rectified Adam (RAdam) optimizer to reduce the $L_1$ loss between the normalized ground-truth flux, $\mathbf{I'}(\mathbf{X}_i)$, and the estimated flux level, $\hat{\mathbf{I}}(\mathbf{X}_i)$, from the hybrid PIML model across $N$ measurements ($i = 1, 2, ..., N$). Equation~\eqref{eqn:optimizer} mathematically represents the fitting process, where $\alpha_t$ is a learning rate at the epoch $t$. The model is trained for 3,000 epochs, with the learning rate initialized at 0.001 and decaying by 10\% every 300 steps.
\begin{equation}
    \theta_{t+1} = \theta_t - \alpha_t \cdot \text{RAdam} \left( \nabla_{\theta} \sum_{i=1}^{N} \frac{\| \mathbf{I'}(\mathbf{X}_i) - \hat{\mathbf{I}}(\mathbf{X}_i; \theta_t) \|_1}{N} \right)
    \label{eqn:optimizer}
\end{equation}

Note that the detected flux sequence itself serves as training data, with all parameters fitted to estimate the source location $\hat{\mathbf{S}}$. This means that fitting the model with the input flux sequence simultaneously infers the radiation source location.

\subsection{Parallel Inference}
\label{section:parallel_inference}
Fitting performance can be improved across multiple trials due to the randomness of the initialization. However, there are two challenges from multiple trials: (i) Longer computation time, (ii) Uncertain performance evaluation when the ground-truth source location is unknown. The process is executed on CPUs for the broad applicability across a range of edge devices.

To address the longer computation time across multiple trials, parallel inference can be implemented using multiprocessing, rather than sequential execution. Twelve models are randomly initialized and executed in parallel. Still, the next challenge is to define the best outcome across multiple models executed. The strategy for selecting the best model is to choose the model with the lowest $L_1$ loss. Although the $L_1$ loss does not directly evaluate the RSL accuracy, it is related to the precision of the reconstructed flux level derived from the estimated source location and other model parameters. The assumption underlying the use of the $L_1$ loss to select the best model is that the model that best fits the measured flux level also performs best for RSL. This assumption will be verified in Section~\ref{section:sim_results}.

\subsection{Continuous Learning}
When the proposed framework is deployed on a real robot, the radiation flux is measured over time. Only the most recent flux measurements within a fixed-size memory buffer are used for RSL. Throughout this paper, this buffer size is referred to as the ``lookback window size".

In real robot deployment, the RSL process is implemented online. However, fitting the model from scratch at each time step is highly inefficient. To address this, the PIML model is first optimized from scratch with a learning rate of 0.005 for 1,500 epochs using the same decaying strategy in Section~\ref{section:optimization}. Subsequently, the model is fine-tuned with a constant learning rate of 0.005 for 50 epochs. This approach is one kind of continuous learning~\cite{liu2017lifelong}. Parallel inference remains integrated throughout the continuous learning process. By leveraging previously optimized parameters and a reduced number of epochs, the proposed RSL framework enables real-time online implementation on the robot. 

Since the robot is unaware of the full size of the environment, the rescale factors ($\text{scale}_x$,  $\text{scale}_y$) are dynamically determined by the adaptively adjusted global bounds, $x$-coordinate, and $y$-coordinate. The global bounds are adaptively updated whenever the current measurement location exceeds the corresponding global min/max values. Subsequently, the previously estimated source location $\hat{\mathbf{S}}$ is re-aligned based on the amount of change of global bounds. To get adjusted trainable tensor values of ($\hat{s}_x$, $\hat{s}_y$), the re-aligned $\hat{\mathbf{S}}$ is normalized by the new rescale factors, and then passed through the logit function~\eqref{eqn:logit}, which is the inverse of the sigmoid.
\begin{equation}
L(p) = \ln\left(\frac{p}{1-p}\right)
\label{eqn:logit}
\end{equation}
In addition, the values of the attenuation tensors should be shifted to be reused. The attenuation tensors $\mathbf{a}$ within the lookback window size $l$ at the timestamp $t$ can be represented as equation~\eqref{eqn:a_t}.
\begin{equation}
    \mathbf{a}^{(t)} = \left[ a_1^{(t)}, a_2^{(t)}, \dots, a_l^{(t)} \right]
    \label{eqn:a_t}
\end{equation}
The $a_i^{(t)}$ estimates whether the $i$-th measured gamma-ray flux is attenuated by obstacles or not at the timestamp $t$. At timestamp $t+1$, the first $k$ radiation measurements overlap with the last $k$ elements of those at the previous time step $t$. Therefore, to reuse the previous attenuation tensors, the last $k$ elements of $\mathbf{a}^{(t)}$ are shifted left by $l-k$ into $\mathbf{a}^{(t+1)}$, and the last $l-k$ values of $\mathbf{a}^{(t+1)}$ are initialized as 0. This process is presented in equation~\ref{eqn:a_t+1}.
\begin{equation}
    \mathbf{a}^{(t+1)} = [ \underbrace{a_{l-k+1}^{(t)}, a_{l-k+2}^{(t)} \dots, a_{l}^{(t)}}_{k \text{ elements}}, \underbrace{0, \dots, 0}_{l-k \text{ elements}} ]
    \label{eqn:a_t+1}
\end{equation}
\section{Results in Simulation}
\label{section:sim_results}
The proposed method has been validated across a range of systematically generated unstructured simulation environments. Gamma-ray flux is simulated using the high-fidelity Monte Carlo particle transport code, OpenMC~\cite{romano2015openmc}. In addition, a random robot path generator has been developed to determine measurement trajectories. The generated path traverses only free space, ensuring collision avoidance with configured obstacles in the OpenMC simulation and maintaining the specified safety margin from the radiation source.

\subsection{Dataset Generation}
\subsubsection{OpenMC Simulation Setup}
\label{subsection:openmc_simulation_setup}
There are three classes of 2D spatial scales for OpenMC simulation: $20\mathrm{m}\times20\mathrm{m}$, $10\mathrm{m}\times10\mathrm{m}$, and $5\mathrm{m}\times5\mathrm{m}$. For each environmental scale, a varying number of obstacles, $N_{\text{obs}}\in\{0, 10, 20, 30, 40, 50\}$, are randomly distributed at non-overlapping locations. Hence, the total number of combinations of 3 scales and 6 $N_{\text{obs}}$ is 18. For each environmental configuration, 2,000 simulation samples are collected, featuring randomized source types and locations, and diverse obstacle materials and geometries. Therefore, the total number of OpenMC samples is 18 × 2,000 = 36,000. 

Eighteen industrial radiation sources are selected for use in the environments: Ir-192, Se-75, Co-60, Cs-137, Cs-134, Co-57, I-131, Y-88, Na-22, Eu-152, Am-241, Ra-226, Tl-208, Ba-133, Zn-65, Mn-54, Cu-64, and Ag-110m. Gamma radiation energies and emission probabilities are referenced from the International Atomic Energy Agency (IAEA) standards~\cite{iaea_gamma_standards,nichols2008evaluation}. Each source is modeled as isotropic, with an activity of 1 MBq. The source is randomly placed within a square centered in the environment space, with a side length equal to half the environment's size. The number of simulated particles is 200,000, with a batch size of 50.

The material of each obstacle is randomly selected among five types: concrete, water, iron, polyethylene, and lead. Each obstacle is randomly chosen to be a circle or a rectangle with the respective dimensions of width, length, and diameter being sampled from $2.5\%$ to $5\%$ of the environment's edge length. The diversity of obstacle materials and geometries results in complex attenuation patterns that are difficult to estimate. The mass densities and elemental compositions of the materials are defined according to the PNNL-15870~\cite{detwiler2021compendium}.

\begin{figure*}[t]
  \centering
  \subfloat[20$\mathrm{m}$ $\times$ 20$\mathrm{m}$]{ 
    \includegraphics[width=0.3\textwidth]{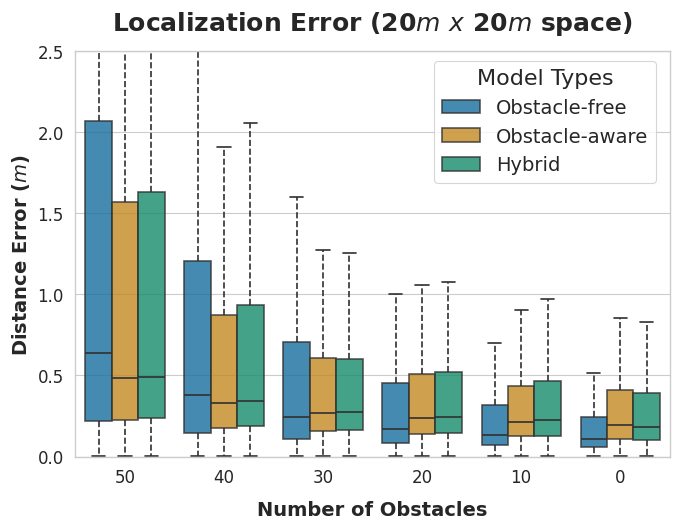}
    \label{fig:d_err_20}
  }\hfill
  \subfloat[10$\mathrm{m}$ $\times$ 10$\mathrm{m}$]{
    \includegraphics[width=0.3\textwidth]{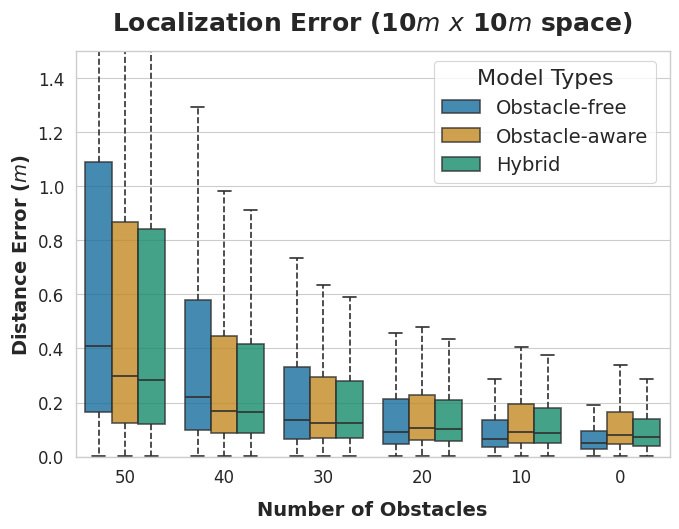}
    \label{fig:d_err_10}
  }\hfill
  \subfloat[5$\mathrm{m}$ $\times$ 5$\mathrm{m}$]{
    \includegraphics[width=0.3\textwidth]{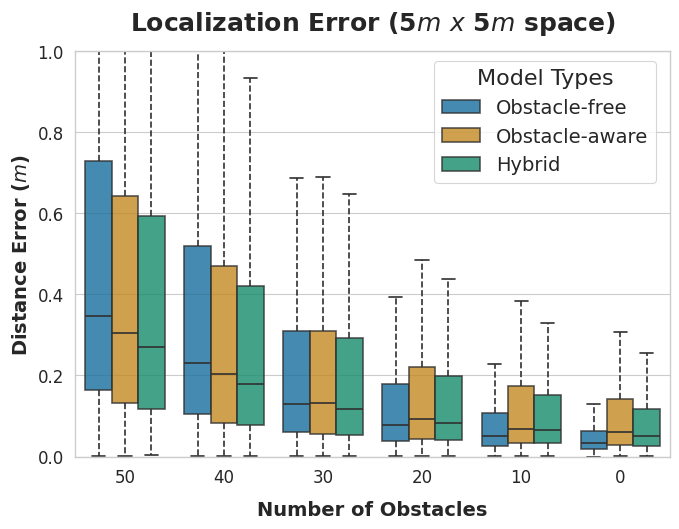}
    \label{fig:d_err_5}
  }
  \caption{Distance error distribution according to the number of obstacles in each environmental scale, (a) $20\mathrm{m}\times20\mathrm{m}$ environment, (b) $10\mathrm{m}\times10\mathrm{m}$ environment, (c) $5\mathrm{m}\times5\mathrm{m}$ environment}
  \label{fig:d_errs}
\end{figure*}

\subsubsection{Arbitrary Robot Path}
\label{section:arbitrary_robot_path}
For each OpenMC simulation sample, three distinct robot paths are randomly generated, yielding a total dataset size of 108,000. The path is drawn from a start point, through two waypoints, to a final end point. The start point, end point, and waypoints are randomly selected in the environment. The A* algorithm is utilized to find a trajectory for each path section~\cite{hart1968formal}. The step size of nodes is $5\%$ of the environment's edge length, and diagonal movement is enabled. The radiation at each node is measured by reading the corresponding flux value from the OpenMC simulation results. The detected flux values are obtained via Gaussian sampling, using the mean and variance for each mesh obtained from the OpenMC simulation.

The generator infinitely re-tries to produce a path until the following three rules are satisfied. First, the generated path is constrained to free space, ensuring no collisions with obstacles. Second, the generated path maintains a minimum safety margin of 1.25$\mathrm{m}$ from the radiation source, thereby ensuring the non-trivial difficulty of the generated RSL task while prioritizing the radiological safety of the robot. Third, the number of measurement nodes along the generated path must range from 30 to 60. Notably, the number of measurements is reduced by a factor between 1/4 and 1/2 relative to the previous PIML-based robotic RSL approach~\cite{son2025physics}. The mean number of nodes is 40.88, and the standard deviation is 8.36. This indicates a right-skewed distribution, as the midpoint of the range 30 to 60 is 45, making the RSL task challenging due to the small number of measurement points.

\begin{figure}[t]
    \centering    
    \includegraphics[width=0.85\columnwidth]{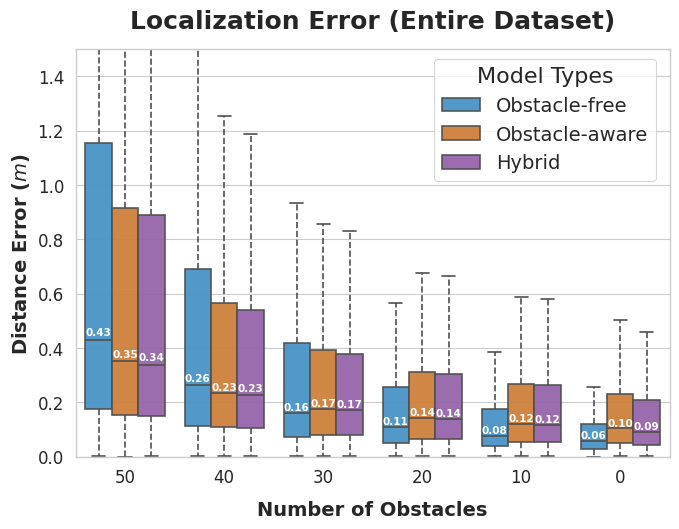}
    \caption{Distance error distribution according to the number of obstacles across all environmental scales}
    \label{fig:d_err_all}
\end{figure}

\subsection{Distance Error}
The primary performance metric is the distance error ($e_{\text{dist}}$), defined as the Euclidean distance between the ground-truth (GT) source location and the estimated source location, as shown in equation~\eqref{eqn:e_dist}.
\begin{equation}
    e_{\text{dist}} = \| \mathbf{S} - \mathbf{\hat{S}} \|_2
    \label{eqn:e_dist}
\end{equation}
\subsubsection{$e_{\text{dist}}$ Distribution}
\begin{figure*}[t]
    \centering
    \includegraphics[width=.95\textwidth]{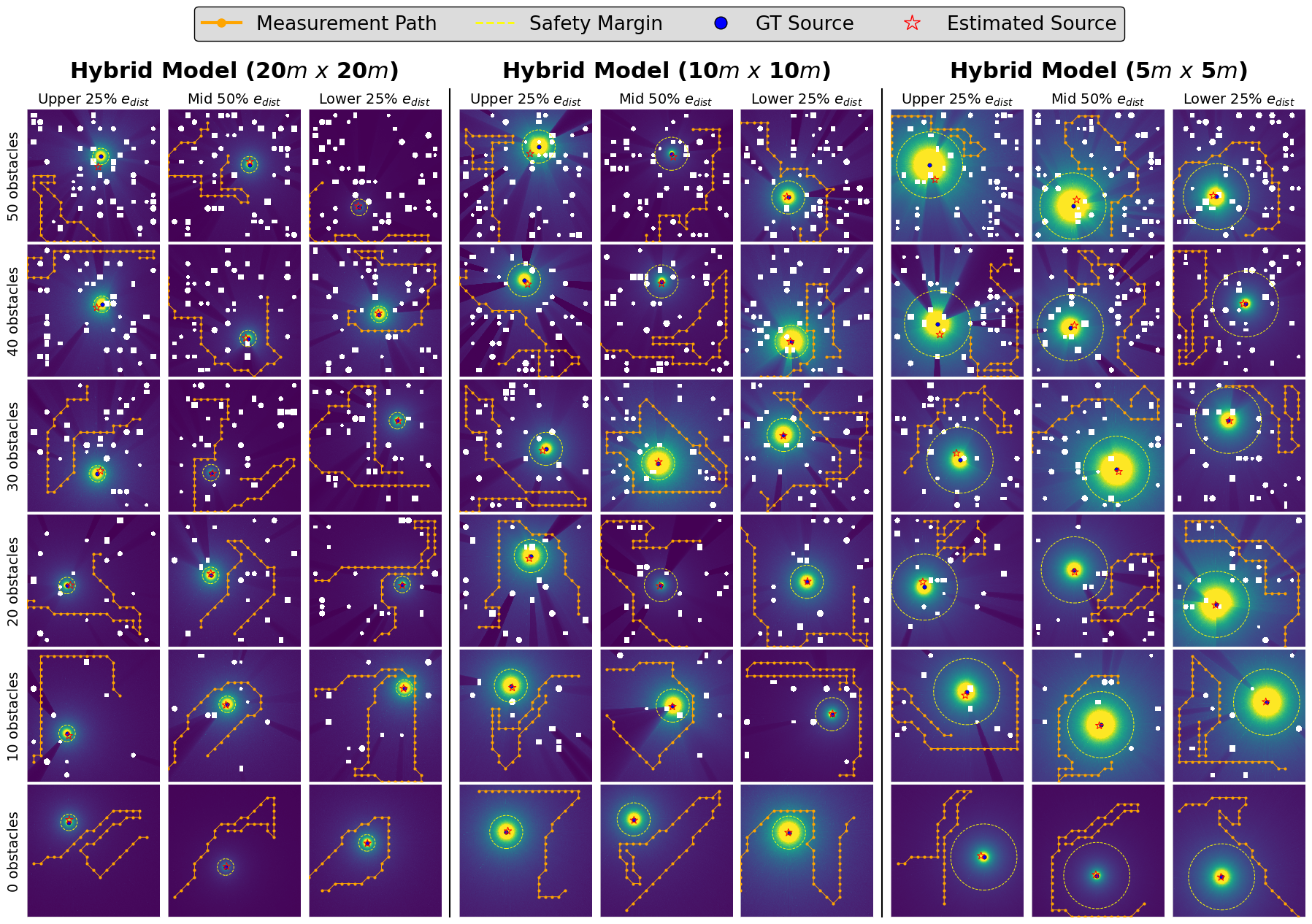}
    \caption{Visual samples of RSL results across multiscale environments with varying obstacle counts}
    \label{fig:rsl_all}
\end{figure*}

Figure~\ref{fig:d_errs} shows the $e_{\text{dist}}$ distribution across 18 environmental setups (scales and the number of obstacles defined in Section~\ref{subsection:openmc_simulation_setup}). The box plots are constructed following Tukey's standard, where whiskers extend to 1.5 times the interquartile range (IQR)~\cite{tukey1977exploratory}. Distance error distributions of 20$\mathrm{m}$$\times$20$\mathrm{m}$, 10$\mathrm{m}$$\times$10$\mathrm{m}$, and 5$\mathrm{m}$$\times$5$\mathrm{m}$ are represented in Fig.~\ref{fig:d_err_20}, ~\ref{fig:d_err_10}, and ~\ref{fig:d_err_5}, respectively. In addition, Fig.~\ref{fig:d_err_all} presents the distance error distributions by number of obstacles across all environmental scales. Figure~\ref{fig:rsl_all} visualizes qualitative RSL results for all 18 setups, representing specific instances that correspond to the 25\%, 50\%, and 75\% distance error levels shown in the box plots of Fig.~\ref{fig:d_errs}. The safety margin around the source is represented by a yellow dashed line, and the generated robot path (an orange line) is shown not to intersect it. White objects are randomized obstacles, and the ground-truth source location and the estimated source location are indicated by a blue dot and a red star, respectively. 

The general trend of the $e_{\text{dist}}$ distribution indicates that the hybrid or obstacle-aware models outperform the obstacle-free model in environments with a large number of obstacles (high-complexity environments), whereas the obstacle-free model performs better in environments with a small number of obstacles (low-complexity environments). This trend is well represented in Fig.~\ref{fig:d_err_all}. It is worth noting that the hybrid model clearly outperforms the others when the scale is small and environmental complexity is high, as shown in Fig.~\ref{fig:d_err_5}. In large-scale, high-complexity environments, the hybrid model achieves the same performance as the obstacle-aware model and still outperforms the obstacle-free model, as shown in Fig.~\ref{fig:d_err_20} and \ref{fig:d_err_10}.

Fitting the obstacle-free model is simpler than fitting the hybrid model because it has fewer parameters. Consequently, it is reasonable to expect that the obstacle-free model outperforms the hybrid model in sparse environments with fewer obstacles. However, as the number of obstacles increases, this performance advantage is reversed, with the hybrid model providing more accurate source localization by accounting for obstacle-induced attenuation. The results illustrated in Fig.~\ref{fig:rsl_all} well align with this intuition that radiation flux measurements along arbitrary robot paths are less affected, or sometimes not affected, by the attenuation when the number of obstacles is 0, 10, or 20 (low-complexity environments). As the number of obstacles exceeds 30 (in high-complexity environments), attenuation increases due to the greater number of obstacles between the robot's path and the radiation source. 

\subsubsection{Mean $e_{\text{dist}}$ Comparison}

Table~\ref{tab:comparison_results} presents a comparison of the mean $e_{\text{dist}}$ across different robotic RSL methods under varying levels of environmental complexity. Additionally, it details the ablation studies of the proposed hybrid model, including component-level analysis (offset tensors and parallel inference) and architecture-level comparisons (obstacle-free and obstacle-aware). It firstly shows the RSL performance across all environments, using the entire dataset. For the detailed analysis, the environment is classified as low- or high-complexity based on the number of obstacles, denoted by $N_{\text{obs}}$. In each category, the best model's results are indicated in bold and underlined. Overall, the proposed hybrid model performs best in total environments, achieving a mean $e_{\text{dist}}$ of $0.53\mathrm{m}$.

\begin{table*}[t]
\centering
\caption{Mean Distance Error ($m$) Comparison across Different Environmental Complexities}
\label{tab:comparison_results}
\setlength{\tabcolsep}{5pt}
\resizebox{\textwidth}{!}{
\begin{tabular}{lcc|c|cc|cc}
\toprule
\multirow{2}{*}{\textbf{Method}} & \multirow{2}{*}{\stackanchor{\textbf{Offset}}{\textbf{Tensors}}} & \multirow{2}{*}{\stackanchor{\textbf{Parallel}}{\textbf{Inference}}} & \textbf{Total Envs} & \multicolumn{2}{c|}{\textbf{Low-Complexity Envs}} & \multicolumn{2}{c}{\textbf{High-Complexity Envs}} \\ \cmidrule(lr){4-4} 
\cmidrule(lr){5-6} \cmidrule(lr){7-8}
& & & $N_{\text{obs}}\in$ ALL & $N_{\text{obs}}=0$ & $N_{\text{obs}}\in\{0, 10, 20\}$ & $N_{\text{obs}}=50$ & $N_{\text{obs}}\in\{30, 40, 50\}$ \\
\midrule
Ours (\textit{Hybrid Model}) & \checkmark & \checkmark & \textbf{\underline{0.53}} & 0.35 & 0.39 & \textbf{\underline{0.89}} & \textbf{\underline{0.67}} \\
Ours (\textit{Hybrid Model}) & \checkmark & $\times$ & 1.75 & 1.49 & 1.57 & 2.11 & 1.94 \\
Ours (\textit{Hybrid Model}) & $\times$ & \checkmark & 3.14 & 2.93 & 3.00 & 3.41 & 3.28 \\
Ours (\textit{Obstacle-free Model}) & \checkmark & \checkmark & 0.54 & \textbf{\underline{0.25}} & \textbf{\underline{0.32}} & 1.06 & 0.78 \\
Ours (\textit{Obstacle-aware Model}) & \checkmark & \checkmark & 0.53 & 0.35 & 0.39 & 0.90 & 0.68 \\ \hdashline
Prev. (\textit{PIML-based, 2025}) \cite{son2025physics}& - & $\times$ & Diverge & Diverge & Diverge & Diverge & Diverge \\
Prev. (\textit{PIML-based, 2025}) \cite{son2025physics}& - & \checkmark & Diverge & Diverge & Diverge & Diverge & Diverge \\
Prev. (\textit{PF, 2025}) \cite{lazna2025localizing}& - & - & 2.07 & 2.01 & 2.03 & 2.16 & 2.11 \\
Prev. (\textit{EKF+ML-EM, 2024}) \cite{nguyen2024radiation}& - & - & 2.43 & 2.15 & 2.24 & 2.82 & 2.62 \\
Prev. (\textit{STE of Dual-Stage, 2024}) \cite{zhu2024dual}& - & - & 4.25 & 2.23 & 3.79 & 4.80 & 4.71 \\
\bottomrule
\end{tabular}}
\end{table*}

For the component-level ablation study of the hybrid model, the table presents two optional components: (i) offset tensors $O=\{o_1, o_2, o_3\}$ and (ii) parallel inference. Omitting one of the two components underscores the importance of leveraging both. Across all environments, it is obviously shown that parallel inference and offset tensors in the presented hybrid PIML model are essential for RSL performance. The performance is notably worse when parallel inference or offset tensors are omitted than when no components are missing.

For the architecture-level ablation study of the hybrid model, which isolates and evaluates only one of the two constituent models, the table also includes the obstacle-free and obstacle-aware models for comparison. The obstacle-free model outperforms others in low-complexity environments. Still, the hybrid model performs best across all environments or only in high-complexity environments. Although the performance gap between the hybrid and obstacle-aware models is small, the hybrid model remains a reasonable best choice. As illustrated in Fig.~\ref{fig:d_err_5}, it distinctly outperforms the obstacle-aware model in high-complexity settings at a small environmental scale. 

Additionally, because the proposed hybrid method is PIML-based, the state-of-the-art previous PIML-based robotic RSL method is included in the comparison~\cite{son2025physics}. In the previous study, the method performed perfectly well using the predefined robot path along four edges of obstacle-free environments with a constant radiation source type. However, given the environmental settings of this paper, the model parameters from the previous study diverge, resulting in undefined loss values (NaN), even with parallel inference or when $N_{\text{obs}}=0$. This indicates that the radiation source type and robot path randomization significantly increase the difficulty of the RSL problem, and the proposed hybrid model performs well on it. Therefore, the proposed method represents a remarkable advancement over the previous study.

Other recent robotic RSL methods are also included in the comparison. First, the particle filter (PF)~\cite{lazna2025localizing} is compared as a baseline method. The control algorithm in Reference~\cite{lazna2025localizing} is unnecessary because the comparison in Table~\ref{tab:comparison_results} is not about path planning; instead, it uses arbitrary measurement trajectories. Also, the attenuation effect is ignored due to the assumption of Reference~\cite{lazna2025localizing}. The RSL performance of PF is worse than that of the proposed hybrid model, with a mean $e_{\text{dist}}$ that is 3.9 times larger in total environments.

Second, the combination of the extended Kalman filter (EKF) and maximum-likelihood expectation-maximization (ML-EM)~\cite{nguyen2024radiation} method is used for comparison. The EKF is used to find a high-probability area where a radiation source may be located, and the ML-EM processes the RSL. Reference~\cite{nguyen2024radiation} explained that characteristics of obstacles are obtained by 3D object detection and corrected with a known blueprint of the environment. However, the comparison in Table~\ref{tab:comparison_results} requires a challenging scenario where obstacle geometries and material properties are completely unknown. Therefore, to include Reference~\cite{nguyen2024radiation} in the comparison, it is assumed that obstacle properties can be fully obtained by 3D object detection. Even though the EKF+ML-EM method uses the entire map with obstacle properties, it performs worse than the proposed hybrid model, even worse than PF, with a mean $e_{\text{dist}}$ of $2.43\mathrm{m}$ in total environments.

Third, the source term estimation (STE) of dual-stage planner~\cite{zhu2024dual} is included in the comparison. STE is the first stage of Reference~\cite{zhu2024dual}, which estimates the radiation source's location from current-flux measurements. The second stage is the ``relocation stage", which moves the robot to the next sub-area. Only the STE is valid for evaluating performance based on arbitrary robot paths and corresponding flux measurements. Without the path planner (second stage), STE shows the worst performance among the benchmarks of Table~\ref{tab:comparison_results}. This implies that arbitrary robot paths with sparse radiation flux measurements make the RSL task significantly more challenging and emphasizes that the proposed hybrid model effectively addresses this challenge.





\subsection{Accuracy}
The accuracy of RSL is measured by the success rate (SR), defined as the proportion of trials in which the radiation source is successfully localized among $N$ total samples. The success of the $i$-th estimation, denoted by $\text{SC}_i$, is quantified using a distance threshold $d_{\text{th}}$. Specifically, as shown in equation~\eqref{eqn:SC_i}, $\text{SC}_i$ is assigned a value of 1 if the distance error of the $i$-th sample, $e_{\text{dist}}^i$, is within $d_{\text{th}}$, indicating a successful localization; otherwise, $\text{SC}_i$ is 0.
\begin{equation}
\text{SC}_i = \begin{cases}
1, & \text{if } e_{\text{dist}}^i \le d_{\text{th}} \\ 
0, & \text{if } e_{\text{dist}}^i > d_{\text{th}} 
\end{cases}
\label{eqn:SC_i}
\end{equation}
Then, the SR is represented as equation~\eqref{eqn:SR}. The unit of SR is percentage (\%).
\begin{equation}
\text{SR (\%)} = \frac{100}{N} \sum_{i=1}^{N} \text{SC}_i
\label{eqn:SR}
\end{equation}

\begin{table}[t]
\centering
\caption{RSL Success Rate (\%) Comparison}
\label{tab:sr_comparison}
\setlength{\tabcolsep}{4pt}
\resizebox{\columnwidth}{!}{
\begin{tabular}{lccc}
\toprule
\multirow{2}{*}{\textbf{Method}} &  \textbf{SR (\%)} & \textbf{SR (\%)} & \textbf{SR (\%)} \\ 
\cmidrule(lr){2-4}
& $d_\text{th}=1.50$ & $d_\text{th}=0.75$ & $d_\text{th}=0.25$ \\
\midrule
Ours (\textit{Hybrid Model}) & \textbf{\underline{90.74}} & \textbf{\underline{84.74}} & \textbf{\underline{63.19}}
\\
Prev. (\textit{PIML-based, 2025}) \cite{son2025physics} & Diverge & Diverge & Diverge \\
Prev. (\textit{PF, 2025}) \cite{lazna2025localizing} & 43.33 & 10.36 & 1.35 \\
Prev. (\textit{EKF+ML-EM, 2024}) \cite{nguyen2024radiation} & 32.09 & 8.86 & 0.98 \\
Prev. (\textit{STE of Dual-Stage, 2024}) \cite{zhu2024dual} & 16.85 & 4.55 & 0.50 \\
\bottomrule
\end{tabular}}
\end{table}

Table~\ref{tab:sr_comparison} shows the comparison of SR across the proposed hybrid model and previous methods that are listed in Table~\ref{tab:comparison_results}. The entire dataset is used for SR evaluation, and the distance threshold, $d_{\text{th}}$, is selected as 1.50$\mathrm{m}$, 0.75$\mathrm{m}$, and 0.25$\mathrm{m}$, respectively. The proposed hybrid model significantly outperforms existing methods across all $d_{\text{th}}$ settings. While the success rate (SR) of previous methods drops sharply as $d_{\text{th}}$ decreases from 1.50$\mathrm{m}$ to 0.75$\mathrm{m}$, and nearly collapses at 0.25$\mathrm{m}$ (below 1.35\% SR), the proposed model maintains high accuracy. Users can expect a 90.74\% success of RSL within 1.50$\mathrm{m}$, maintaining a high reliability of 84.74\% (only 6\% reduction) within 0.75$\mathrm{m}$, and still offers a 63.19\% chance of success even at the high-precision, 0.25$\mathrm{m}$ threshold.

\subsection{Parallel Inference}

\subsubsection{Computation Time}
\label{section:computation_time}
\begin{figure}[t]
  \centering
  \subfloat[Operational Complexity]{ 
    \includegraphics[width=0.47\columnwidth]{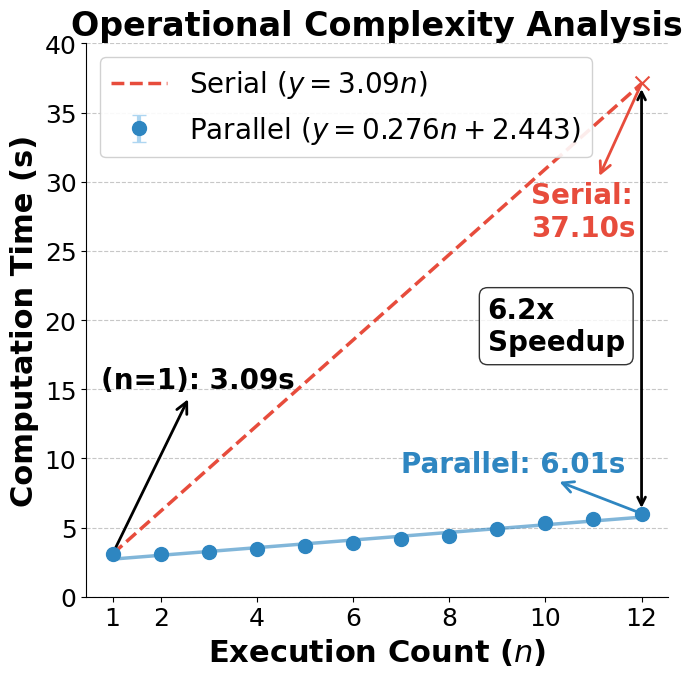}

    \label{fig:operational_complexity}
  }
  \hfill
  \subfloat[Speedup Factor]{ 
    \includegraphics[width=0.47\columnwidth]{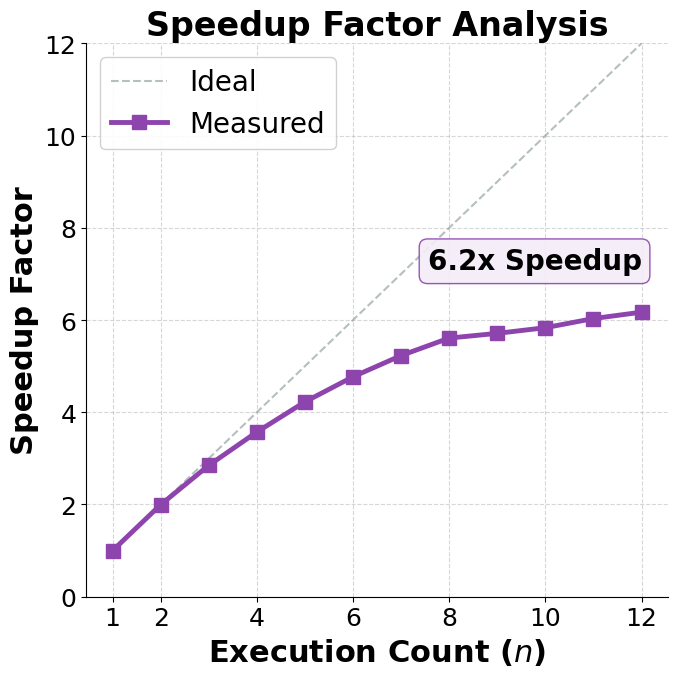}
    \label{fig:speedup_factor}
  }
    \caption{Computation time analysis of parallel inference in the hybrid model}
    \label{fig:parallel_analysis}
\end{figure}

As mentioned in Section~\ref{section:parallel_inference}, parallel inference can address the long computation time across multiple executions. However, the actual speedup achieved by parallelization is inherently limited by Amdahl's law and Gunther's Universal Scalability Law (USL)~\cite{rodgers1985improvements,gunther1993simple}. Figure~\ref{fig:parallel_analysis} shows the computation time analysis of the hybrid model with parallel inference. The figure illustrates the operational complexity and the speedup factor as functions of the number of executions. An Intel Core i7-13700 CPU with 16 GB of RAM is used to measure computation times. Also, 10 samples are randomly selected for each $N_{\text{obs}}$ and environment scale (a total of 180 samples), and the average computation time is calculated.

Figure~\ref{fig:operational_complexity} presents the operational complexity for serial and parallel inference, showing the advantages of parallel inference in terms of computation time. Computation time for a single execution is 3.09 seconds with the hybrid model. It is expected that the computation time increases proportionally with the number of executions in the serial inference, yielding 37.10 seconds for 12 executions. For the same number of executions, parallel inference takes 6.01 seconds, which is 6.2 times (speedup factor) faster than serial computing.  The regressed line slope drops from 3.09 to 0.276 when transitioning from serial to parallel inference. The computation time for 12 parallel processes scales by a factor of 1.94 relative to a single-process execution.

Figure~\ref{fig:speedup_factor} illustrates the speedup factor analysis of parallel inference. In an ideal scenario, the speedup factor equals the number of parallel inference processes. Parallel inference performs ideally when the number of executions is less than 4. However, once the execution count exceeds 4, processes begin to contend for shared resources and become saturated at 8. Therefore, for real-world robot deployment, it is practical to utilize the maximum number of workers only for the initial estimation, while limiting parallel workers to 3 for subsequent continuous learning to avoid performance degradation.

\begin{figure*}[t]
  \centering
  \subfloat[20$\mathrm{m}$ $\times$ 20$\mathrm{m}$]{ 
    \includegraphics[width=0.3\textwidth]{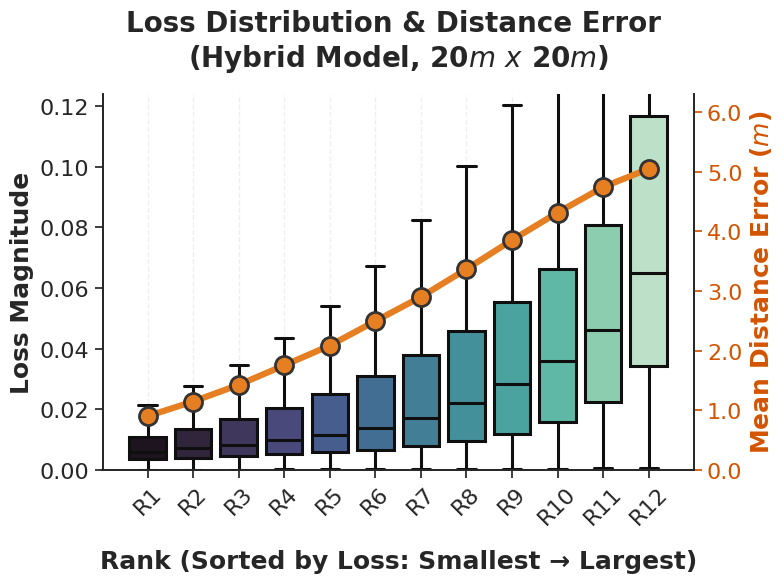}

    \label{fig:loss_err_distribution_20}
  }\hfill
  \subfloat[10$\mathrm{m}$ $\times$ 10$\mathrm{m}$]{ 
    \includegraphics[width=0.3\textwidth]{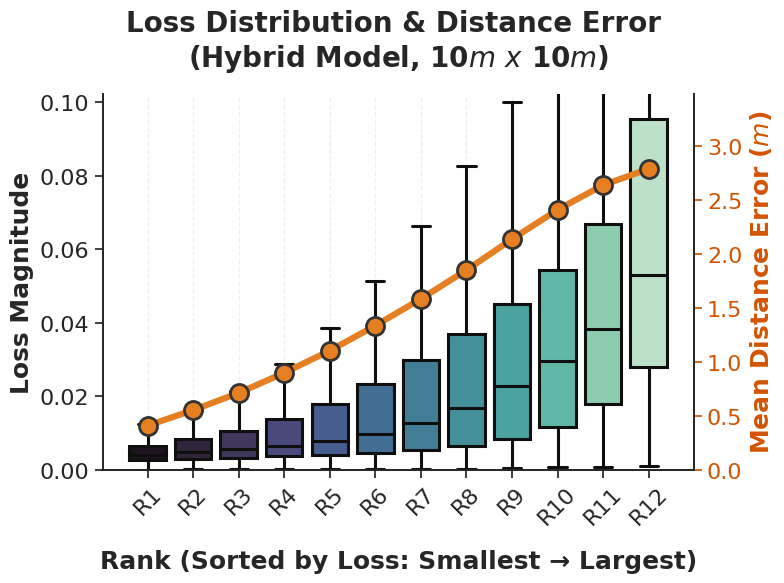}
    \label{fig:loss_err_distribution_10}
  }\hfill
  \subfloat[5$\mathrm{m}$ $\times$ 5$\mathrm{m}$]{
    \includegraphics[width=0.3\textwidth]{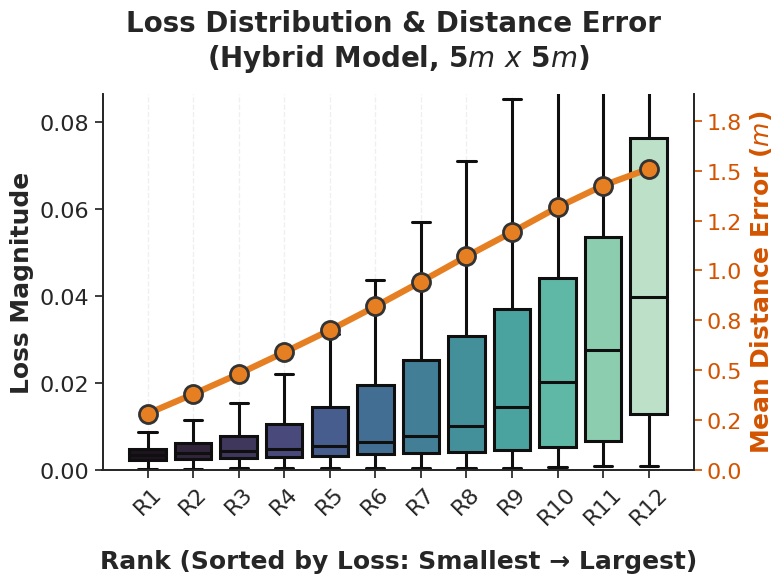}
    \label{fig:loss_err_distribution_5}
  }
  \caption{$L_1$ loss distribution and mean distance error with parallel inference of the hybrid model across multiscale environments}
  \label{fig:loss_err_distribution}
\end{figure*}

\subsubsection{Model Loss and Distance Error}
\label{sec:model_loss_and_distance_err}
Once multiple models are executed in parallel, the next step is to choose the best model that produces the most precise source location. The strategy for selecting the best outcome is to compare $L_1$ losses. However, while model parameters are optimized to minimize the $L_1$ loss between $\mathbf{I'}(\mathbf{X})$ and $\mathbf{\hat{I}}(\mathbf{X})$, minimizing flux reconstruction error does not inherently guarantee precise source localization. Figure~\ref{fig:loss_err_distribution} shows the correlation between $L_1$ loss and $e_{\text{dist}}$. For each environmental scale, $L_1$ losses from the 12 parallel inferences are ranked in ascending order, and the corresponding mean distance error is obtained at each rank. The figure shows a clear correlation: the lower the $L_1$ loss, the more precise the RSL. Therefore, selecting the minimum $L_1$ loss model from the parallel inference outputs is empirically justified.

\begin{figure*}[t]
  \centering
  \subfloat[Minimum Achievable $e_{\text{dist}}$]{ 
    \includegraphics[width=0.3\textwidth]{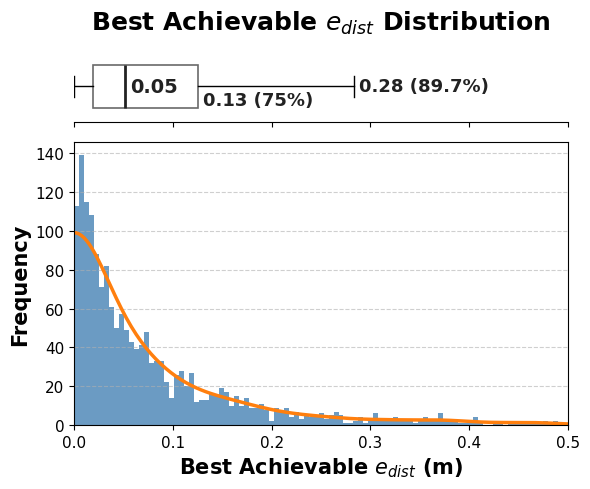}

    \label{fig:best_ach}
  }\hfill
  \subfloat[Optimality Gap]{ 
    \includegraphics[width=0.3\textwidth]{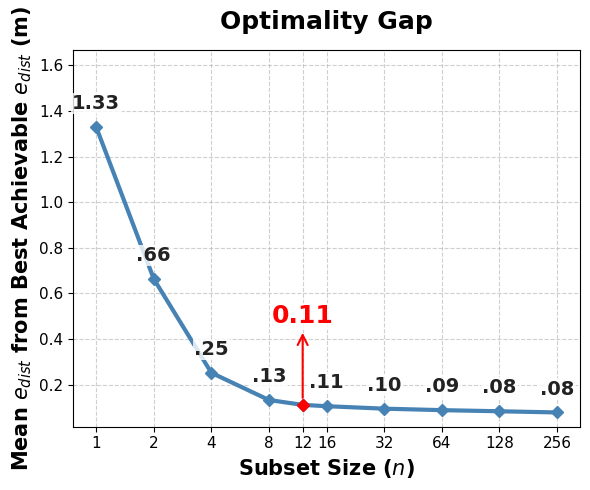}
    \label{fig:opt_gap}
  }\hfill
  \subfloat[Statistical Stability]{
    \includegraphics[width=0.3\textwidth]{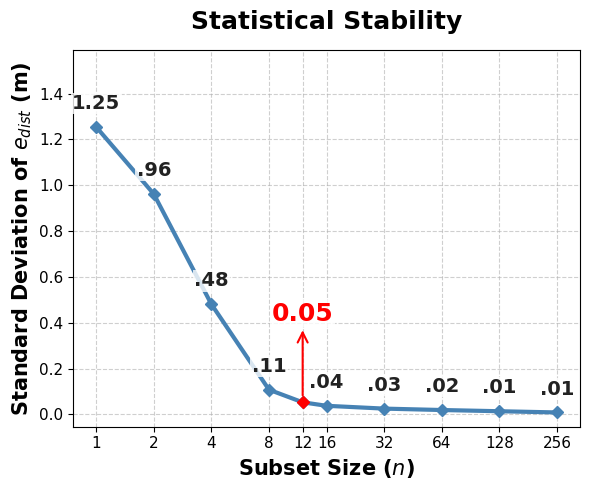}
    \label{fig:stat_stability}
  }
  \caption{Robustness of parallel inference and lowest $L_1$ loss model selection against random parameter initialization, (a) Minimum achievable distance error of the hybrid model, (b) Optimality gap between the mean distance error and minimum achievable distance error, (c) Statistical stability demonstrated by standard deviation}
  \label{fig:robustness}
\end{figure*}

\subsubsection{Robustness}
Although the positive correlation between the model loss and the distance error is presented in Section~\ref{sec:model_loss_and_distance_err}, the robustness of a random parameter initialization with a certain number of parallel executions is still unclear and needs to be justified. To verify this, 100 map-path pairs are randomly sampled from a pool of 6,000 data points for each of the 18 environmental setups, yielding a total of 1,800 samples in the evaluation set. Then, the proposed RSL processing is repeated 500 times for each sample, with random initialization, to demonstrate its robustness.

The focus here is not on the consistency of RSL performance across 500 randomly initialized trials, but rather on how the parallel inference technique mitigates uncertainty from random initializations and presents robust output. The initial step in this evaluation is to identify the model's fundamental limits in terms of the minimum achievable $e_{\text{dist}}$. The minimum achievable $e_{\text{dist}}$ of each sample can be quantified by selecting the best iteration, which yields the minimum $e_{\text{dist}}$, across 500 randomly initialized trials. Figure~\ref{fig:best_ach} illustrates the distribution of the minimum achievable $e_{\text{dist}}$ for all 1,800 samples. The distribution is extremely right-skewed, with a median of 0.05$\mathrm{m}$. This indicates that the proposed model's fundamental limits are sufficiently close to the optimal, zero error.

To show the robustness of the parallel inference, the specific subset size, $n_{\text{subset}}$, is determined. Then, for each sample, $n_{\text{subset}}$ trials are randomly chosen from 500 trials, and $e_{\text{dist}}$ of the one that shows the minimum $L_1$ loss is selected as an output. This process is the same as the best model selection strategy of the proposed RSL method. The random sampling is conducted 500 times for each map-path pair. Consequently, a total of 500 distinct parallel inference results ($n_{\text{subset}}$ executions), each with distinct random parameter initializations, are generated for the identical map-path pair sample across all 1,800 samples.

Figure~\ref{fig:opt_gap} shows the optimality gap, which is the difference between the average $e_{\text{dist}}$ of 500 distinct parallel inferences and the minimum achievable $e_{\text{dist}}$ in each sample. Also, Fig.~\ref{fig:stat_stability} shows the statistical stability, which is the standard deviation of $e_{\text{dist}}$ of 500 distinct parallel inferences in each sample. For both figures, the median value across 1,800 samples is plotted as a function of $n_{\text{subset}}$. While the distance error remains high and unstable in the absence of parallel inference ($n_{\text{subset}}=1$), increasing $n_{\text{subset}}$ effectively reduces the optimality gap and stabilizes $e_{\text{dist}}$ from random parameter initializations. The performance effectively saturates beyond $n_{\text{subset}}=12$, with both the optimality gap and standard deviation diminishing to near-zero values. Regarding practical computational resources, selecting 12 executions for parallel inference can be justified to obtain accurate and consistent results, yielding an expected $e_{\text{dist}}$ offset of 0.11$\mathrm{m}$ from ideal performance, with a standard deviation of 0.05$\mathrm{m}$.

\section{Experimental Validation}
\label{section:experimental_validation}
The proposed hybrid model for robotic RSL has been validated through two types of experiments using a real radiation detector and a robot. For the first experiment (Experiment A), gamma-ray count-per-second (CPS or $\gamma\,\mathrm{s}^{-1}$) was collected manually at each lattice point in the environment. The arbitrary robot paths are generated from these measurements, and the robustness and precision of RSL are validated. For the second experiment (Experiment B), the radiation detector was integrated into the robot system. While collecting gamma-ray CPS every second, the robot navigated through spontaneous teleoperation in three different scenarios. The RSL was conducted with continuous learning and validated under a range of lookback window sizes.

\subsection{Experiment A: Manual Lattice Detections}

\begin{figure*}[htbp]
  \centering
  \subfloat[Lattice Experiment Setup]{ 
  \raisebox{6mm}{
    \includegraphics[width=0.25\textwidth]{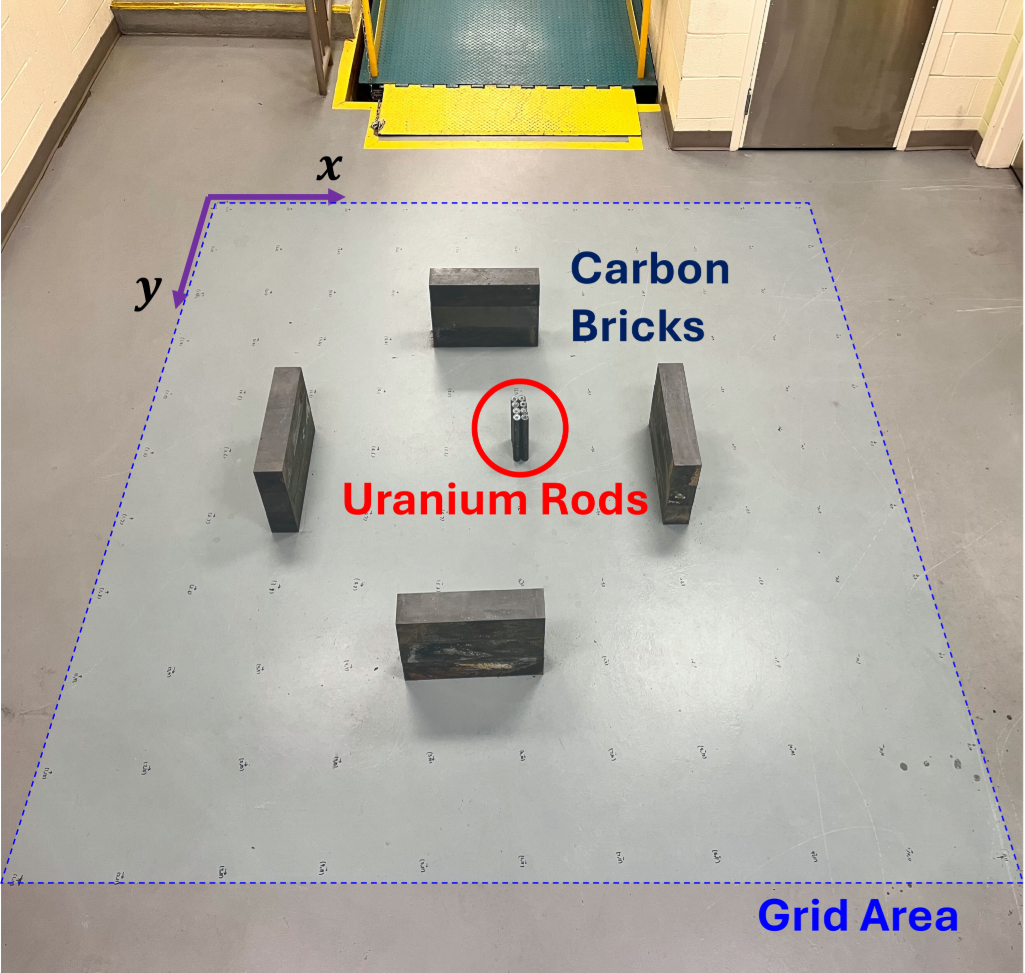}
  }
    \label{fig:lattice_experiment_env}
  }\hfill
  \subfloat[RSL with Edge Path]{
    \includegraphics[width=0.345\textwidth]{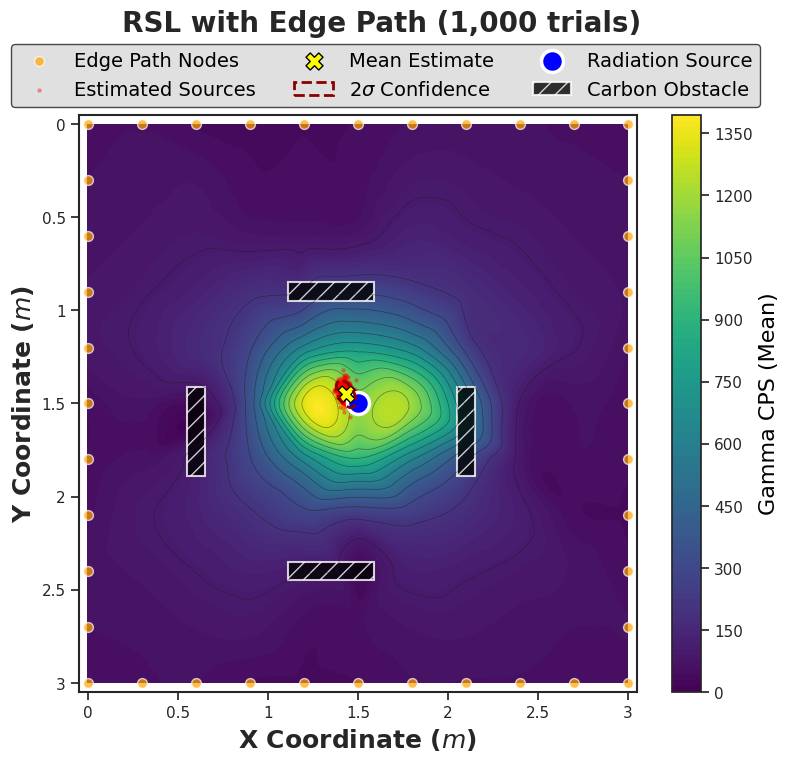}
    \label{fig:lattice_experiment_edge}
  }\hfill
  \subfloat[RSL with Arbitrary Paths (21 nodes)]{ 
    \includegraphics[width=0.35\textwidth]{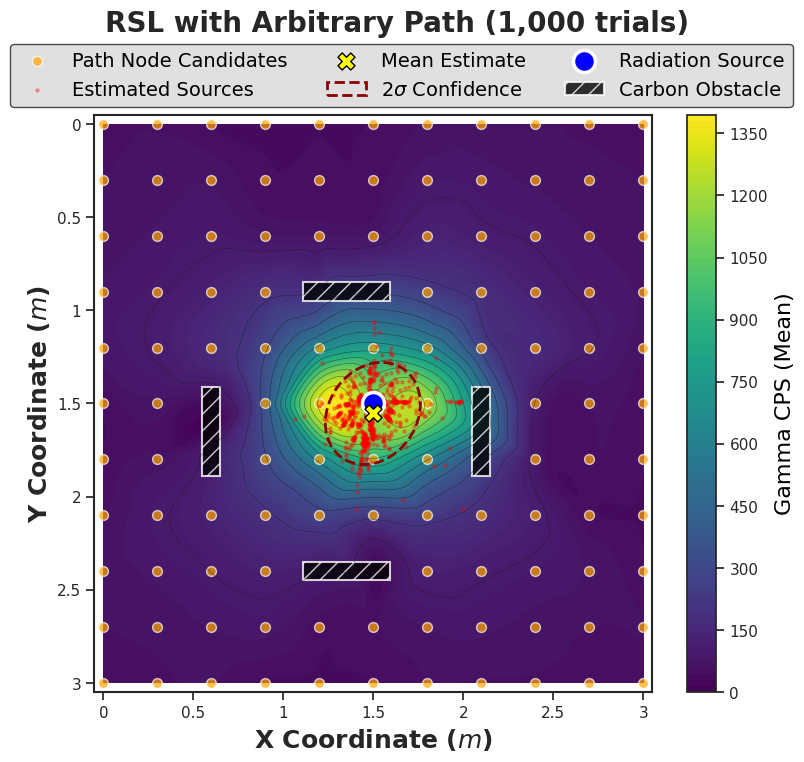}
    \label{fig:lattice_experiment_arbitrary}
  }
  \caption{Experimental setup and localization performance for Experiment A: Manual Lattice Detections}
  \label{fig:lattice_experiment}
\end{figure*}

\subsubsection{Dataset}
The same experimental dataset and setup reported by Tan \textit{et al.} are used to verify the proposed method. \cite{tan2025rapid}. In the experiment, a hand-carried detector, RadEye SPRD-ER~\cite{mayfield2019thermo}, was placed on the ground to collect the gamma-ray CPS at a constant height. The environment is a $3.0\mathrm{m}\times3.0\mathrm{m}$ square room, which is discretized into an $11\times11$ lattice. The CPS were measured at all lattice intersections at every $0.3\mathrm{m}$. The physical layout is illustrated in Fig.~\ref{fig:lattice_experiment_env}, and the lattice measurement points are represented in Fig.~\ref{fig:lattice_experiment_edge} and \ref{fig:lattice_experiment_arbitrary} with orange dots. As shown in Fig \ref{fig:lattice_experiment_env}, four carbon wall segments serving as obstacles with in-plane size $0.48\times 0.10\mathrm{m}$ and height $0.30\mathrm{m}$ are arranged around the eight cylindrical uranium rods placed in the center, (1.5$\mathrm{m}$, 1.5$\mathrm{m}$) point. Notably, carbon and uranium have not been included in simulations of Section~\ref{section:sim_results}. During data collection, five CPS values were measured repeatedly at each lattice point.


\subsubsection{Edge Path}
\label{section:edge_path}
The first experiment set with manual lattice detections aims to evaluate the proposed method's robustness against sensor noise. Five repeated measurements at each point are varied by sensor noise, which represents the inherent uncertainty in radiation detection. To evaluate the robustness of the proposed method against sensor noise, the measurement path is fixed to the edge path, and CPS values are randomly selected from five measurements at each node. The RSL with randomly selected CPS values was repeated 1,000 times. For practical purposes, background radiation is not subtracted because it is difficult to distinguish it accurately in real-life radioactive environments. 

RSL results with the edge path using manual lattice detections are shown in Fig.~\ref{fig:lattice_experiment_edge}. For 1,000 trials, small red dots are estimated source locations, and the mean distance error is 0.088$\mathrm{m}$. Also, the average location of estimated radiation sources is (1.43$\mathrm{m}$, 1.45$\mathrm{m}$), which is 0.083$\mathrm{m}$ away from the actual source location. The uncertainty of the output is evaluated with 2$\sigma$ confidence. The squared Mahalanobis distance follows a chi-squared distribution with two degrees of freedom~\cite{mahalanobis2018generalized}. Consequently, the 2$\sigma$ ellipse boundary represents a cumulative probability of 86.5\%, shown by the dark red dashed line. The major and minor axes of the 2$\sigma$ confidence ellipse are 0.16$\mathrm{m}$ and 0.085$\mathrm{m}$, respectively, yielding a geometric mean radius of $0.058\mathrm{m}$. It is concluded that 86.5\% of estimated sources are located within a 0.058$\mathrm{m}$ geometric mean radius of a center point that is offset by 0.083$\mathrm{m}$ from the actual source. It sufficiently confirms the robustness of the proposed RSL method against sensor noise.


\subsubsection{Arbitrary Paths}
The second experiment set evaluates RSL precision from arbitrary robot paths. All the measurement strategies, except for robot paths, are the same as in the first experiment set, which utilized an edge path. For each trial, an arbitrary robot path with 21 measurement nodes was generated using the same method in Section~\ref{section:arbitrary_robot_path}, including start point, end point, and two waypoints. The 21 nodes equal the total number of nodes encountered when moving along the edges from one vertex of the space to the diagonally opposite vertex. With an arbitrary robot path using randomly selected CPS values, more general precision and robustness can be validated than with the fixed-edge path. 

Figure~\ref{fig:lattice_experiment_arbitrary} illustrates RSL results with arbitrary robot paths using manual lattice detections. The mean distance error is 0.17$\mathrm{m}$, and the average location of estimated sources is (1.50$\mathrm{m}$, 1.55$\mathrm{m}$), which is 0.053$\mathrm{m}$ away from the actual source location. In addition, the major and minor axes of the 2$\sigma$ confidence ellipse are 0.58$\mathrm{m}$ and 0.48$\mathrm{m}$, respectively. This indicates that 86.5\% of estimated sources are within a 0.26$\mathrm{m}$ geometric mean radius, with the center situated 0.053$\mathrm{m}$ away from the GT source. These values fall within the acceptable range, demonstrating the practical feasibility of the proposed model.

\subsection{Experiment B: Online Robotic RSL}


The first step for the online robotic RSL is to develop the robotic radiation detection system. The integration strategy mirrors the prior study by Son \textit{et al}~\cite{son2025robot,zhang2025icra}. The RadEye SPRD-ER radiation detector is interfaced with a Windows-based single-board computer (SBC), and this detection system is integrated onto a Unitree Go2 EDU quadruped robot. Each second, the detector updates a CPS and publishes a value to Redis. The RSL system simultaneously subscribes to the ROS2 topic to receive the current robot location and to the Redis key to receive the current CPS value, and maps them at each timestamp~\cite{macenski2022robot}. The data collection process runs asynchronously with the RSL process. Consistent with Experiment A, background radiation is not measured and subtracted for practical reasons.

The RSL performance is validated in three distinct cases. The first two (Scenario 1 and Scenario 2) are conducted in small indoor laboratory spaces. Scenario 1 is an open environment, whereas Scenario 2 is more confined. In both cases, obstacles are manually located at arbitrary positions. The last case (Scenario 3) is conducted in an obviously larger space than Scenarios 1 and 2 and features more complex settings. The environment in Scenario 3 is already cluttered with complex ambient objects; therefore, no additional obstacles are manually added.

For Scenarios 1 and 2, eleven bundled uranium rods are positioned within three carbon bricks, four concrete bricks, one lead brick, and other ambient objects already in the environment. These objects are arbitrarily placed without any specific intent or predefined arrangement. For Scenario 3, a horizontal array of fourteen uranium rods is positioned as a radiation source. No obstacles are manually added to the space because the environment is already complex with ambient objects such as dozens of concrete bricks, a carbon brick tower, plastic drums, paper boxes, chairs, a desk, forklifts, and messy cables. It is worth noting again that neither uranium nor such ambient objects have been included in the simulations of Section~\ref{section:sim_results}.

\begin{figure}[t]
    \centering
    \includegraphics[width=.85\columnwidth]{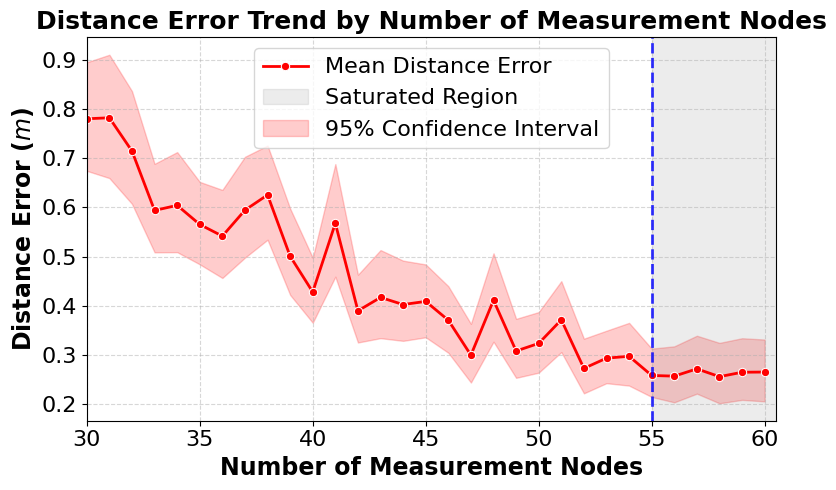}
    \caption{Distance error trend according to the number of measurement nodes in simulation (500 samples per node count)}
    \label{fig:distance_error_by_nodes}
\end{figure}

The online robotic RSL process is executed based on a fixed number of the most recent data points, limited by a predefined lookback window size. One insight from the simulation-based results guides the selection of an appropriate lookback window size. Figure~\ref{fig:distance_error_by_nodes} illustrates the mean distance error with the 95\% confidence according to the number of measurement nodes, from 30 to 60, in simulation-based results. To prevent data imbalance, 500 samples are randomly selected for each measurement node count. As the number of measurement nodes increases, the mean distance error decreases due to richer input information. However, the performance gains saturate beyond a node count of 55. From the data-driven insight, a node count exceeding 55 does not improve model performance. Therefore, it is empirically justified to set the lookback window size to 55.

During the online robotic RSL, the initial estimation is performed with the maximum number of workers, and subsequent continuous learning is performed with three workers, as discussed in Section~\ref{section:computation_time}. The edge board of the Go2 robot is an NVIDIA Jetson Orin Nano with a 6-core CPU. Therefore, the number of workers for the initial estimation is set to 6. At the determined lookback window size of 55, the average computation time on the NVIDIA Jetson Orin Nano is 13.01 seconds for the initial estimation and 0.76 seconds for subsequent updates via continuous learning, which is lower than the radiation sensor's temporal resolution of 1 second and thus feasible for online deployment.

\begin{figure}[t]
  \centering
  \subfloat[Scenario 1]{ 
    \includegraphics[width=.85\columnwidth]{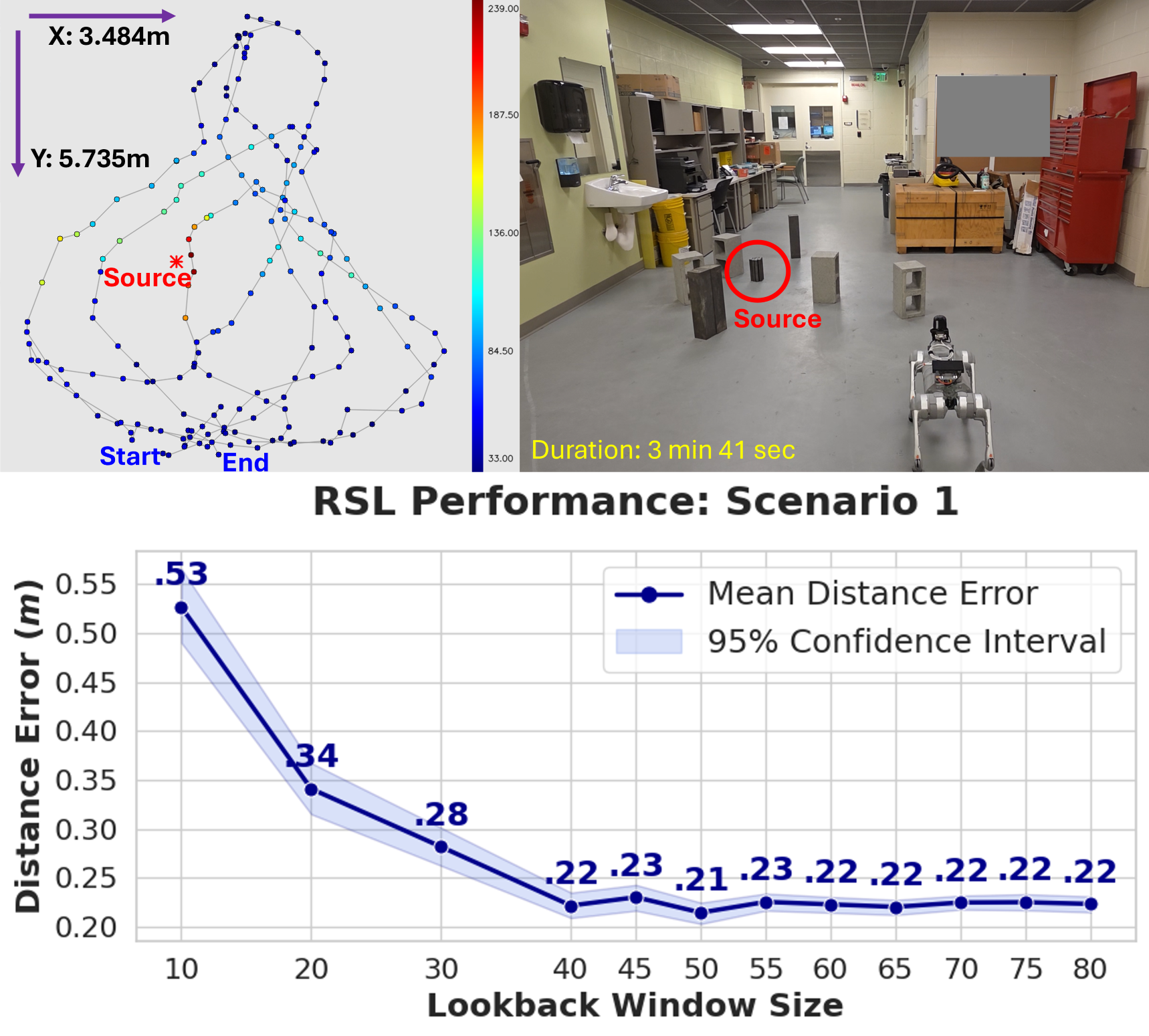}
    \label{fig:result_s1}
  }\hfill
  \subfloat[Scenario 2]{
    \includegraphics[width=.85\columnwidth]{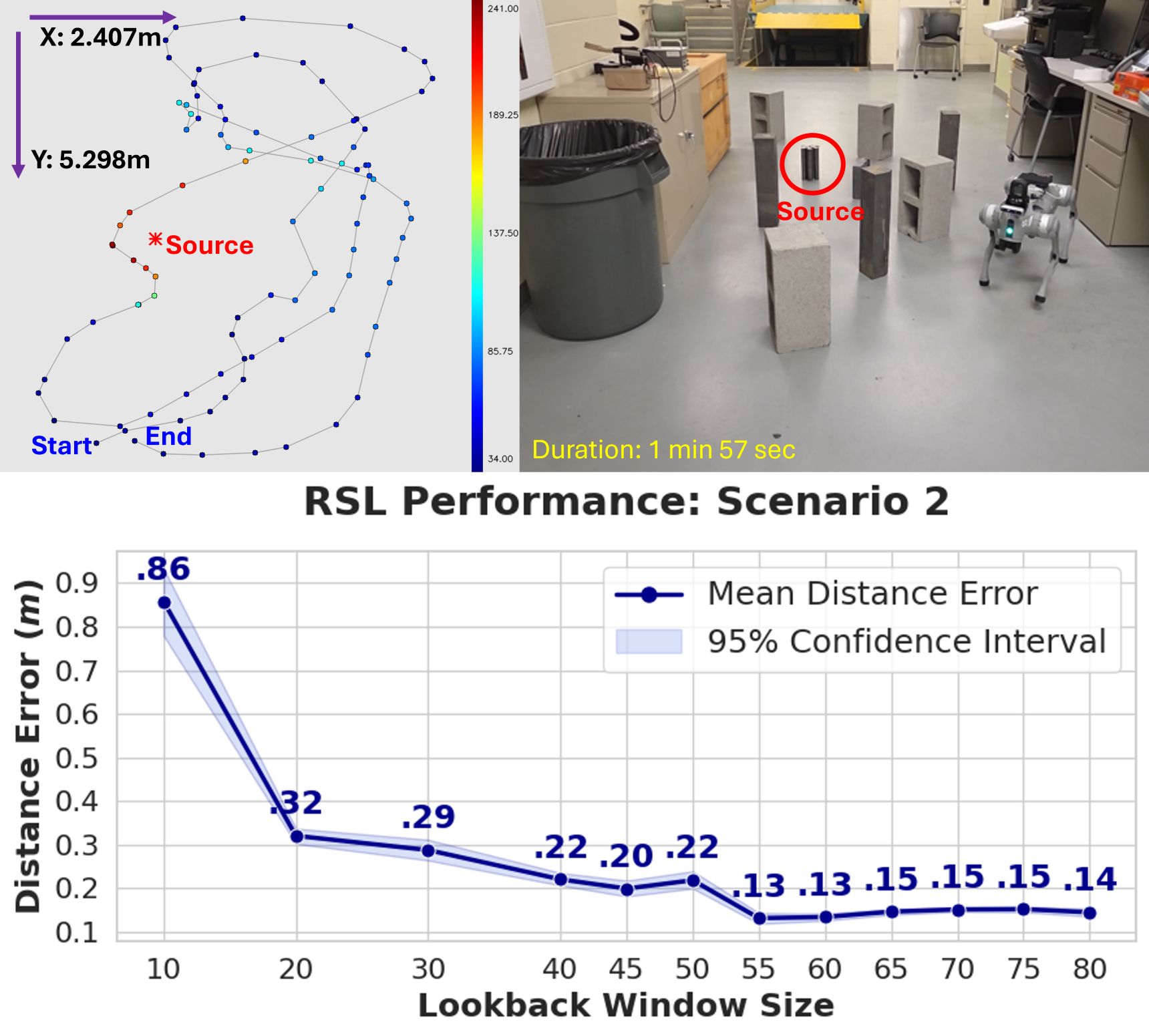}
    \label{fig:result_s2}
  }
  \caption{Experimental setup and localization performance for Experiment B: Online Robotic RSL, (a) Entire robot trajectory, setup picture, and mean distance error of Scenario 1, (3.484$\mathrm{m}$ $\times$ 5.735$\mathrm{m}$, 193 Nodes, Total Duration: 3 min 41 sec), (b) Entire robot trajectory, setup picture, and mean distance error of Scenario 2, (2.407$\mathrm{m}$ $\times$ 5.298$\mathrm{m}$, 98 Nodes, Total Duration: 1 min 57 sec)}
  \label{fig:full_experiment}
\end{figure}

Figure~\ref{fig:full_experiment} shows the actual setup of Experiment B in Scenarios 1 and 2, as well as the entire robot trajectory and the distance error trend according to the lookback window size. The robot was teleoperated by a human in an arbitrary manner, traversing the environment without a specific intent. If multiple radiation CPS measurements are collected at the same location, those are merged into a single measurement by averaging values. In Scenario 1, a total of 193 radiation CPS measurements were collected over a mission duration of 3 minutes and 41 seconds, with the entire trajectory spanning an area of $3.484\,\mathrm{m} \times 5.735\,\mathrm{m}$. In Scenario 2, a total of 98 radiation CPS measurements were collected over a mission duration of 1 minute and 57 seconds, with the entire trajectory spanning an area of $2.407\,\mathrm{m} \times 5.298\,\mathrm{m}$. The performance saturates at lookback window sizes of 40 and 55 in Scenarios 1 and 2, respectively. This saturation trend is consistent with the data-driven saturation point at 55, as determined with Fig.~\ref{fig:distance_error_by_nodes}. Notably, in Scenario 1, small fluctuations occur within the lookback window size range of 40 to 50, followed by full stabilization at 55. The mean distance errors are $0.23\mathrm{m}$ and $0.13\mathrm{m}$ for lookback window size 55 in Scenarios 1 and 2, respectively. These error values align with the $e_{\text{dist}}$ range in $5\mathrm{m}\times5\mathrm{m}$ OpenMC simulation environments.

\begin{figure}[t]
    \centering
    \includegraphics[width=.85\columnwidth]{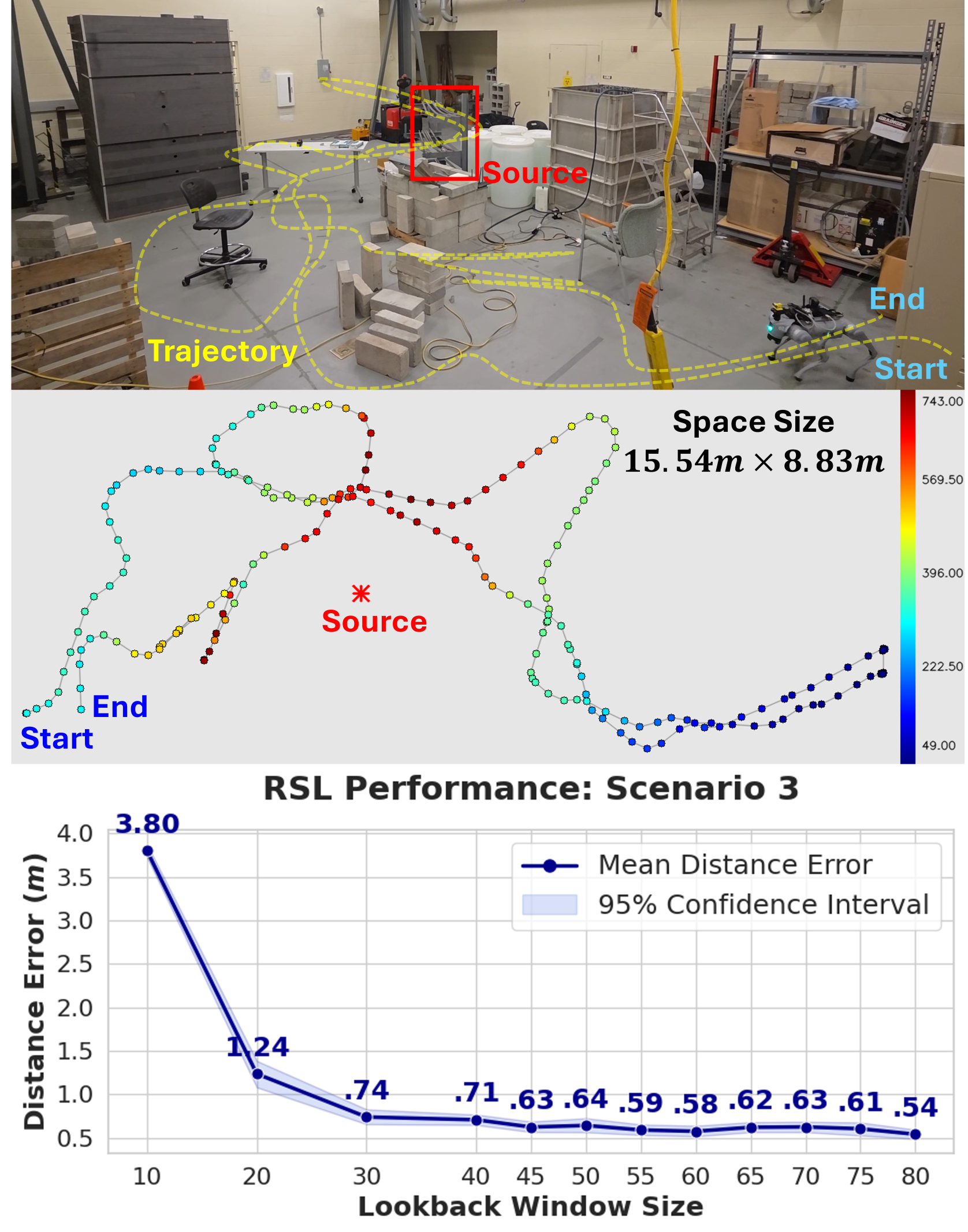}
    \caption{Experimental setup, robot trajectory, and radiation source localization performance for Experiment B: Online Robotic RSL (Scenario 3, 15.54$\mathrm{m}$ $\times$ 8.83$\mathrm{m}$ space, 207 Nodes, Total Duration: 3 min 34 sec)}
    \label{fig:result_s3}
\end{figure}

Figure~\ref{fig:result_s3} illustrates the result of Scenario 3 by showing the actual experimental setup, the entire robot trajectory, and the distance error trend according to the lookback window size. Similar to Scenarios 1 and 2, the robot was teleoperated by a human without special intent, and multiple CPS measurements collected at the same robot location are averaged into a single measurement. The test in Scenario 3 was conducted in the obstacle-cluttered rectangle-shaped room, which is much larger, $15.54\,\mathrm{m} \times 8.83\,\mathrm{m}$, than the areas in Scenarios 1 and 2. The robot's trajectory spans the reachable area of the room. Also, unlike Scenarios 1 and 2, longer uranium rods are laid horizontally across the shelves of a multi-tiered rack, resulting in a stronger anisotropic feature of the radiation source. A total of 207 radiation CPS measurements were collected over a mission duration of 3 minutes and 34 seconds. The saturation trend aligns with the data-driven trend shown in Fig.~\ref{fig:distance_error_by_nodes}. Based on the saturation point in Fig.~\ref{fig:distance_error_by_nodes}, the lookback window size of 55 is selected, resulting in a mean distance error of $0.59\mathrm{m}$. This value aligns with $e_{\text{dist}}$ range of the OpenMC simulation for $20\mathrm{m}\times20\mathrm{m}$ high-complexity environments.

\section{Discussions}
\label{section:discussion}
\subsection{Comprehensive Discussions}
\subsubsection{Radiation Exposure}
The proposed RSL framework is not explicitly designed to avoid high radiation fields, but it provides advanced radiation perception that can be flexibly utilized. While users can actively plan the robot's trajectory using the estimated source location to avoid high radiation exposure, hardware reliability remains crucial for the practical deployment of a robot in highly radioactive environments. Based on Boston Dynamics' white paper, the radiation resilience of the Spot robot was intensively tested at Los Alamos National Laboratory (LANL), demonstrating strong survivability in high-energy gamma fields~\cite{bonn2021radiation}. It shows that the Spot robot can withstand 413 rem of gamma radiation without system failure, an amount equivalent to 82 years of a worker's annual dose limit. In addition, West~\textit{et al.} evaluated the total ionizing dose (TID) tolerance of the Up Core single-board computer (SBC) to ensure reliable operation in high-gamma facilities~\cite{west2022radiation}. The on-board power management circuitry started to malfunction after $111.1 \pm 5.5 \mathrm{Gy}$, which represents a sufficiently high level of radiation tolerance for electronics. They argued that the operational lifetime can even be effectively extended by tungsten shielding or by adjusting the driving voltage. These studies have successfully shown the feasibility of operating robots in highly radioactive environments.

\subsubsection{From Simulation to Real-World}
The proposed RSL framework performed well even in real experimental configurations that fall outside the scope of randomized OpenMC simulations. For example, the OpenMC simulation was implemented in 2D, while the real world is 3D. Also, carbon was not included among the randomly generated obstacles, and uranium was not included among the simulated radiation sources. In addition, the radiation source was modeled as isotropic in the simulation, while bundled uranium rods in real experiments are anisotropic, as shown in the flux distribution of Fig.~\ref{fig:lattice_experiment}. This implies that intensive validation in simulation environments can be a key strategy for addressing data scarcity and the challenges of restricted-access environments. More detailed and realistic synthetic data generation will enhance the model, including 3D objects with well-defined material combinations, time-variant dynamic scenarios in both indoor and outdoor environments, and anisotropic radiation sources. 

\subsubsection{External Factors for Detection Uncertainty}
External factors introduce additional detection noise beyond the inherent Poisson noise. First, the noise arises from the attenuation by the robot frame. The radiation detector is attached to the head of the Go2 robot dog. Therefore, depending on the source location and the robot's orientation, the robot's body frame itself shields gamma radiation. Additionally, if the radiation source is placed below the body height, most of the gamma-ray flux passes through the robot's frame. This increases the detection noise and makes the RSL more challenging. Second, the short dwell time for radiation detection increases the measurement uncertainty. A robot moves continuously, so the number of gamma particles detected per second, which is the SPRD-ER detector's temporal resolution, is not measured at a stationary location. When the robot's target mission allows, this uncertainty can be mitigated by pausing the robot for at least 1 second to collect sufficient CPS at the stationary location.

\begin{figure}[t]
  \centering
  \subfloat[Continuous Learning]{ 
    \includegraphics[width=.85\columnwidth]{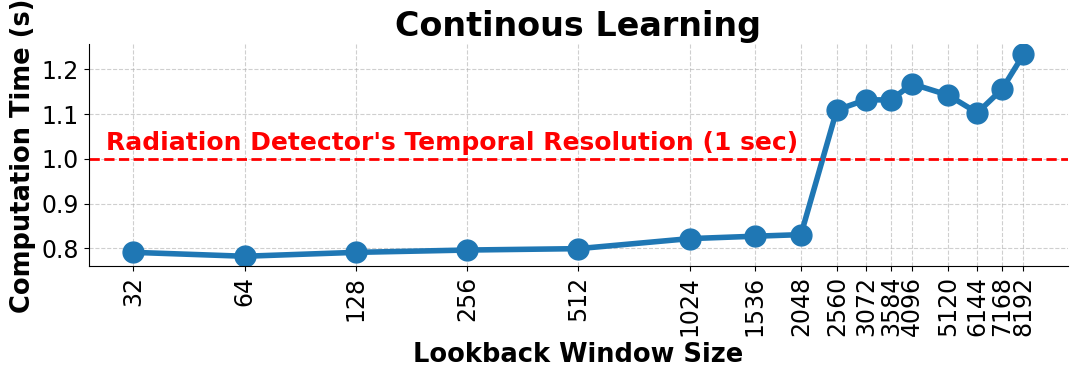}
    \label{fig:jetson_cont_lr}
  }\hfill
  \subfloat[Initial Estimation]{
    \includegraphics[width=.85\columnwidth]{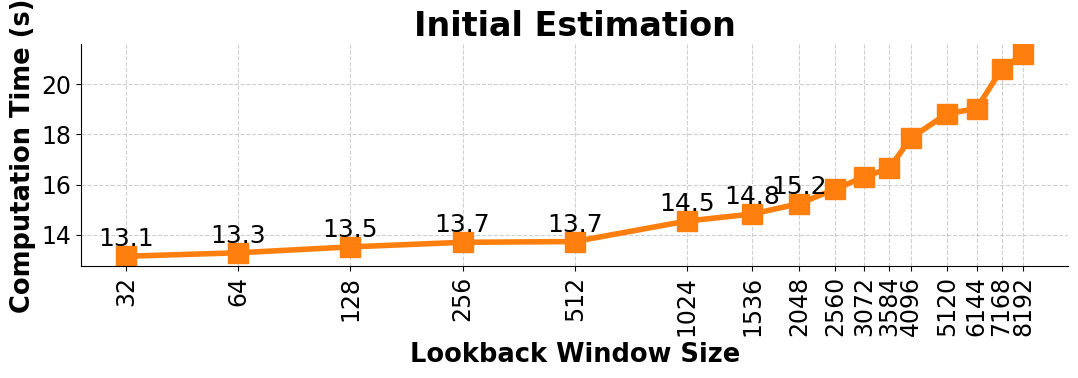}
    \label{fig:jetson_initial_est}
  }
  \caption{RSL Computation time according to the lookback window size using Jetson Orin Nano}
  \label{fig:jetson_computation}
\end{figure}

\subsubsection{Feasible Lookback Window Size}
\label{sec:feasible_lookback_window_size}
Based on data-driven intuition, the lookback window size 55 was determined in online robotic RSL. However, when computational resources are a critical constraint for the robot's target mission, selecting a smaller lookback window size than 55 is preferable. In Fig.~\ref{fig:full_experiment}, with a lookback window size between 20 and 40, mean distance errors range from $0.22\mathrm{m}$ to $0.34\mathrm{m}$ for Scenario 1, and from $0.22\mathrm{m}$ to $0.32\mathrm{m}$ for Scenarios 2. Although these errors are not minimal, the range is acceptable to utilize the corresponding lookback window size. Relative to the total space, the area of the error circle, with a radius of the mean distance error, ranges from 0.76\% to 1.82\% in Scenario 1, and from 1.19\% to 2.52\% in Scenario 2. In the same manner, as shown in Fig.~\ref{fig:result_s3}, the ratio of the error circle to the total space ranges from 1.15\% to 3.51\% in Scenario 3, for lookback window sizes between 20 and 40. These numbers are acceptable to be practically deployed.

From another perspective, it is worth discussing the maximum feasible lookback window size for the online RSL. This is valuable for the larger-scale applications. The total computation time must be shorter than the SPRD-ER detector's temporal resolution (1 second) to ensure real-time continuous learning and avoid data loss of the measured CPS. To determine the maximum feasible lookback window size, the computation time on the Unitree Go2 EDU's edge board was measured for various window sizes.

Figure~\ref{fig:jetson_computation} illustrates the scalability of the lookback window size with respect to the computation time of continuous learning and initial estimation on the Jetson Orin Nano. Results are averaged over 100 samples for continuous learning and 10 samples for the initial estimation. As shown in Fig.~\ref{fig:jetson_cont_lr}, in the continuous learning process, the computation time remains nearly constant until the lookback window size reaches 2,048, but surges above 1 second when the window size exceeds 2,048. It can justify that the maximum feasible lookback window size is 2,048. However, regarding the initial estimation in Fig.~\ref{fig:jetson_initial_est}, the computation time notably increases when the lookback window size exceeds 512. Therefore, 512 can be the maximum lookback window size if timely performance is critical to the robot's target mission, saving approximately 1.5 seconds of the initial estimation time.

\subsubsection{Operational Time}
Not only the computation time for continuous learning and initial estimation, but also the time required to collect the initial CPS array to execute the model, are critical factors in the practical operation time required to obtain the first online RSL result from the actual robotic radiation detection system. The total time to get the first RSL result is combined with two factors that heavily depend on the lookback window size: (i) initial CPS data collection time to fill the input data, and (ii) initial RSL estimation time. CPS data is collected every second, and the expected initial estimation time is presented in Fig.~\ref{fig:jetson_computation}. For example, if the lookback window size is set to 55, it takes 55 seconds to retrieve the CPS array for the initial model execution and 13.2 seconds to obtain the initial RSL estimation result. Totally, it takes 68.2 seconds to obtain the first RSL result with a lookback window size of 55; thereafter, the RSL result is updated every second through continuous learning. Likewise, the practical operation time to get the first RSL result is 45.1 seconds with a lookback window size of 32, and 34 minutes 23.2 seconds with a lookback window size of 2048. Therefore, although 2048 is the maximum lookback window size solely in terms of computation time, it significantly increases operational runtime, so users should carefully select the lookback window size based on the robot's target mission.

\subsubsection{Spatial Scalability}
The proposed RSL method can be spatially scaled up from small to large spaces. As explained in Section~\ref{sec:piml_model_design}, a trainable tensor, $\hat{s}$, is passed through the sigmoid function and rescaled by multiplying scale factors, $(\text{scale}_x, \text{scale}_y)$. Scale factors can be dynamically selected based on the region of interest to localize a radiation source. Although the spatial scalability of the proposed RSL method is validated across various sizes of OpenMC simulation environments and real-world setups, one fundamental challenge is that the distance error increases with larger scale factors. This trend is shown in Fig.~\ref{fig:d_errs}. The reason for this trend is that the magnitude of the gradients from the sigmoid function $\sigma(\hat{s})$ is proportionally scaled up by the scale factors, causing the optimization updates to be too large for precise convergence. Therefore, the relative distance error with respect to environment size is expected to remain acceptably small, but the absolute distance error is predicted to be higher in larger environments. To address this challenge, a mechanism that adaptively adjusts the key hyperparameters, especially learning rate, based on scale factors can be developed in future research to achieve finer convergence and control the absolute distance error.

\subsubsection{Dimensional Scalability}
The scope of the proposed RSL method is focused on ground robot applications, identifying the 2D location of a radiation source. Although the real world is 3-dimensional, the height difference between the detector and the radiation source is relatively small on the 2D ground plane, and can be effectively handled by offset tensors of the proposed hybrid model. Therefore, 2D location estimation remains practical and does not restrict the applicability to ground-robot systems.

Still, the model can be naturally extended to 3D, which is a potential direction for future research. In 3D space, the $i$-th measurement location and ground-truth source location should be redefined as $\mathbf{X_i} = (x_i, y_i, z_i) \in \mathbb{R}^3$ and $\mathbf{S} = (S_x, S_y, S_z) \in \mathbb{R}^3$, respectively. Also, the estimated source location $\mathbf{\hat{S}}$ is redefined by adding the $z$-coordinate.
Consequently, as shown in equations~\eqref{eqn:distance_3d} and~\eqref{eqn:offset_3D}, the estimated distance $\hat{d_i}$ and offset tensors $O$ are calculated in the 3D space. The remaining processes follow the same procedure as the 2D RSL.
\begin{equation}
    \hat{d_i} = \sqrt{(x_i - \hat{S_x})^2 + (y_i - \hat{S_y})^2 + (z_i - \hat{S_z})^2} - \operatorname{Avg}(O)
    \label{eqn:distance_3d}
\end{equation}
\begin{equation}
    O = \{o_1, o_2, o_3\} = \text{MLP}(\sigma(\hat{s_x}), \sigma(\hat{s_y}), \sigma(\hat{s_z}))
    \label{eqn:offset_3D}
\end{equation}

\begin{figure}[t]
    \centering
    \includegraphics[width=.7\columnwidth]{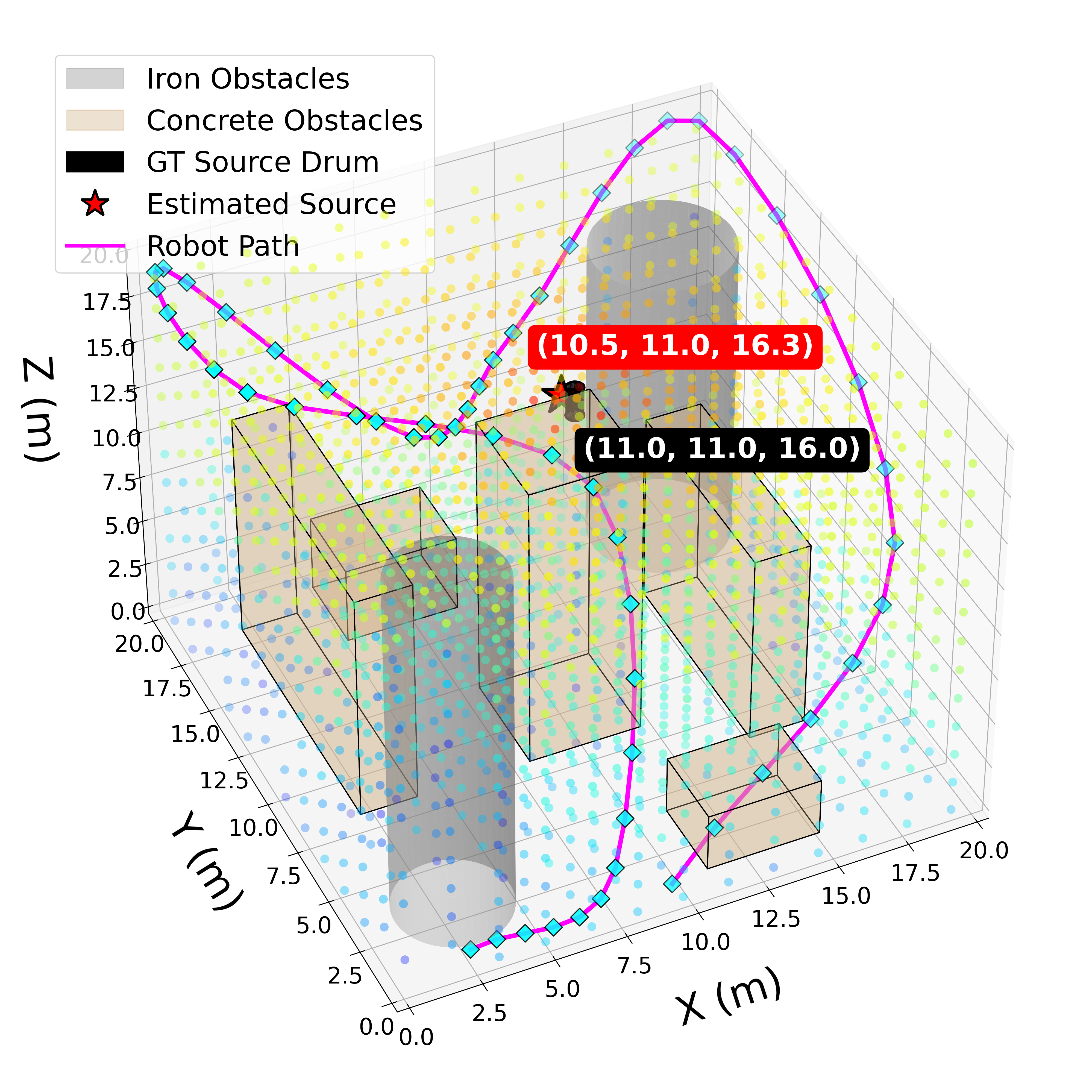}
    \caption{Feasibility demonstration of 3D RSL using an aerial robot}
    \label{fig:3d_rsl} 
\end{figure}

Figure~\ref{fig:3d_rsl} demonstrates the feasibility of extending the proposed RSL framework to 3D RSL. In this demonstration, an aerial robot flies around the environment to collect radiation data. Radiation flux distribution is simulated using OpenMC with the $20\mathrm{m}\times20\mathrm{m}\times20\mathrm{m}$ environment. Multiple concrete and iron obstacles are placed, and the radiation source is located inside the iron drum above the concrete monolith at $(11.0\mathrm{m}, 11.0\mathrm{m}, 16.0\mathrm{m})$. Using the extended hybrid model for 3D RSL, the estimated source location is $(10.5\mathrm{m}, 11.0\mathrm{m}, 16.3\mathrm{m})$ with 55 measurement nodes, indicating a distance error of $0.58\mathrm{m}$. While the current performance is acceptable, 3D RSL introduces additional challenges: (i) the robot's trajectory needs to span multiple altitudes to minimize estimation uncertainty, and (ii) properly tuning the $\text{scale}_z$ parameter is critical for 3D RSL performance. More rigorous validation is needed to provide the precision, robustness, and practical applicability of the extended 3D RSL framework in future research.

\subsubsection{Extensibility to Multiple Radiation Sources}
Although the proposed RSL framework targets a single radiation source, it can be extended to estimate multiple source locations. A major challenge in handling multiple radiation sources is identifying the unknown source count. Based on the limited measurements, the RSL framework should estimate the number of sources, accounting for the possibility of their absence. One straightforward strategy to determine the absence of sources is to develop a heuristic classifier based on the variance of measurement values. For example, if the variance of flux values along the path remains below a specific threshold for a sufficient period, it can be decided that only background radiation exists. Consequently, the classifier determines that there is no distinct radiation source in the environment.

The additional classifier can also determine the number of sources when sources are present. The previous study~\cite{son2025physics} demonstrated the potential to develop a source-count classifier for multi-source cases, but it used only fixed-size inputs and a predefined robot path. Using a source-count classifier for source-existing cases is much more challenging than binary classification of source-absence, due to inconsistent input data shapes and high uncertainties arising from arbitrary robot paths and obstacle-induced attenuation. 

The proposed RSL process, combining parallel inference with minimum-loss model selection, enables the simultaneous identification of the number and locations of multiple radiation sources. To achieve this, the hybrid model for multiple sources should be defined, enabling all single- and multi-source models to run in parallel. Then, the estimated location tensors $\mathbf{\hat{S}}$ of the model with the minimum $L_1$ loss are selected as the final result. To design the hybrid model with multiple radiation sources, tensors of the right-hand side in equation~\eqref{eqn:hybrid} will be individually allocated and fitted for each radiation source, and outputs from each term are summed to calculate the superposed estimated flux $\hat{\mathbf{I}}(\mathbf{X}_i)$, as represented in equation~\eqref{eqn:hybrid_multi}. 
\begin{align}
    \hat{\mathbf{I}}(\mathbf{X}_i) &= \sum_{k=1}^{\text{count}} \left( \sigma(w_k) \cdot \hat{\mathbf{I}}_{\text{free}}(\mathbf{X}_{i,k}) + (1 - \sigma(w_k)) \cdot \hat{\mathbf{I}}_{\text{obs}}(\mathbf{X}_{i,k}) \right) \label{eqn:hybrid_multi} \\
    i &= 1, 2, \dots, N \nonumber
\end{align}

\begin{figure}[t]
  \centering
  \subfloat[Success]{ 
    \includegraphics[width=.85\columnwidth]{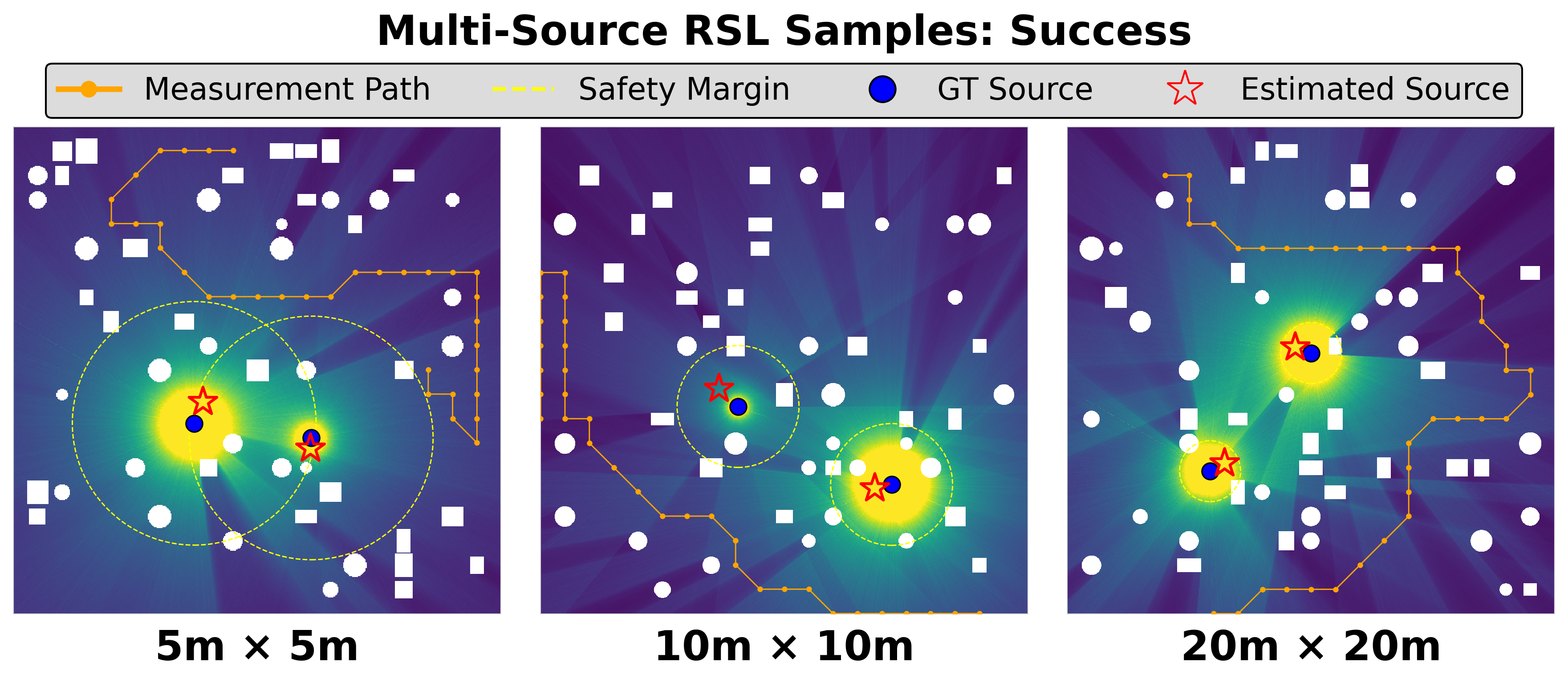}
    \label{fig:multi_rsl_success}
  }\hfill
  \subfloat[Failure (Highly Proximal)]{
    \includegraphics[width=.85\columnwidth]{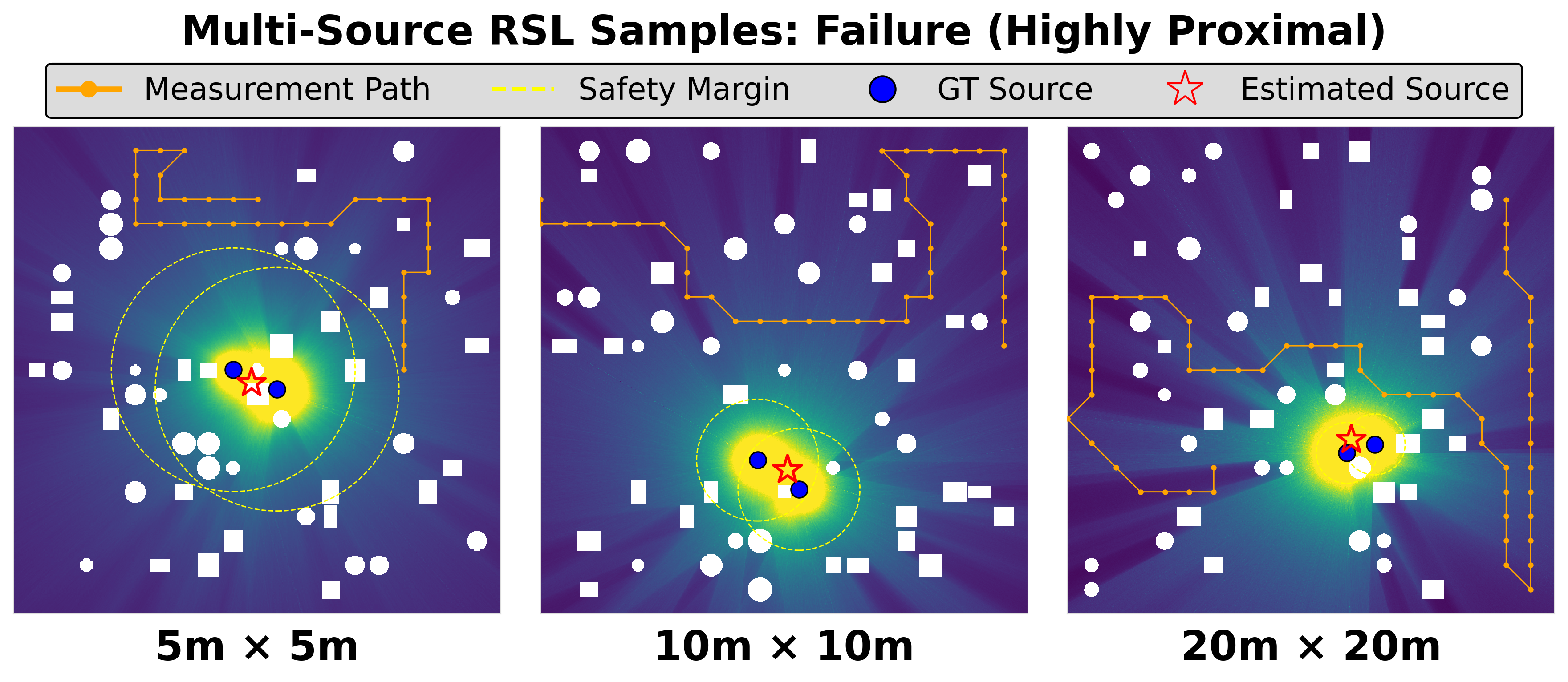}
    \label{fig:multi_rsl_failure_highly_proximal}
  }\hfill
  \subfloat[Failure (Imbalanced Emission Rates)]{
    \includegraphics[width=.85\columnwidth]{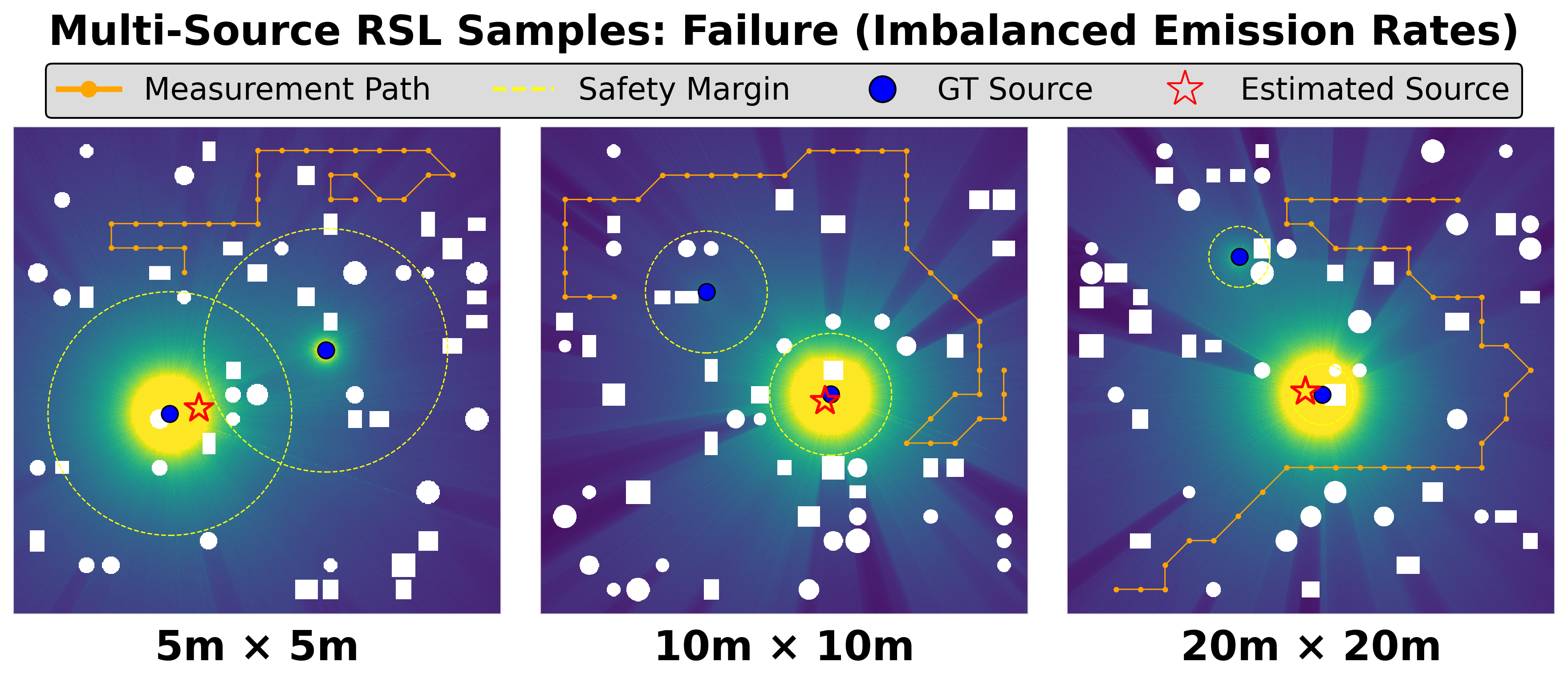}
    \label{fig:multi_rsl_failure_imbalanced_strengths}
  }
  \caption{Qualitative demonstrations of multi-source RSL in high-complexity environments with 50 obstacles, (a) Successful cases, (b) Failed cases due to highly proximal source locations, (c) Failed cases due to imbalanced emission rates}
  \label{fig:multi_rsl}
\end{figure}

The preliminary results demonstrating the potential of the multi-source RSL are presented. It uses 1,800 OpenMC samples for the single-source and multi-source (two-source) cases, respectively. To collect 1,800 data points, 100 samples are randomly collected for each of the 18 environmental setups defined in Section~\ref{subsection:openmc_simulation_setup}. Then, three arbitrary robot paths are generated for each OpenMC sample, resulting in a total of 10,800 data points. Twelve parallel models are executed for single-source and multi-source cases, respectively, for a total of 24 trials. The Receiver Operating Characteristic (ROC) Area Under the Curve (AUC) for source-count classification is 0.7106. In addition, leveraging the error function for multiple sources from Reference~\cite{newaz2016uav}, the total average $e_{\text{dist}}$ is $0.92\mathrm{m}$. The average $e_{\text{dist}}$ decreases to $0.74\mathrm{m}$ for correctly classified samples. These results indicate the potential of the extended hybrid model to simultaneously estimate both the number and locations of multiple sources.

Figure~\ref{fig:multi_rsl} qualitatively represents the multi-source RSL results in high-complexity environments with 50 obstacles. As illustrated in Fig.~\ref{fig:multi_rsl_success}, the model successfully separates multiple radiation sources, handling arbitrary paths and obstacle-induced attenuation. However, when the distance between the two sources is small (as shown in Fig.~\ref{fig:multi_rsl_failure_highly_proximal}) or their emission rates are highly imbalanced (as shown in Fig.~\ref{fig:multi_rsl_failure_imbalanced_strengths}), the combined flux distribution resembles that of a single source, making it difficult for the model to separate the multiple sources. In future research, the method for multi-source RSL needs to be enhanced to address failures, and the performance metric itself may need to be redefined for these edge cases.

\subsection{Limitations \& Future Research Directions} 
\subsubsection{Limitations}
The proposed RSL framework uses only a single ground robot to estimate the 2D location of the radiation source. Another limitation is that it is currently restricted to single-source localization and does not account for multi-source scenarios or source-absence classification. Additionally, experiments with a real radiation source, detector, and robot were conducted only in lab-scale indoor environments, and no adaptive mechanism was introduced to actively determine the lookback window size in the online RSL.

\subsubsection{Future Research Directions}
The proposed method can be extended to operate with heterogeneous multi-robot teams, including aerial robots, in 3D environments. In addition, the loss-comparison strategy can be applied to select the best PIML model capable of handling multiple radiation sources. For the online RSL with continuous learning, the method can be verified in large-scale environments using an adaptive method to select the proper lookback window size. A possible approach to adaptively adjust the lookback window size is to use the variance of the estimated source location in the online RSL, based on the robot's speed or the measured flux gradient. Also, beyond the robotic RSL, the PIML-based approach can be extended to autonomous radiation mapping and reconstruction~\cite{adams2025behavioral,tan2025rapid}.

\section{Conclusion}
\label{section:conclusion}
This study proposes a physics-guided robotic radiation-source localization framework based on parallel inference of a PIML model. The design of the trainable tensors is inspired by the physical principles of gamma radiation, and the fitted source-location tensors are retrieved to localize the radiation source. The proposed method performs in unstructured environments with diverse obstacles and radiation source types, collecting sparse gamma-ray flux along an arbitrary robot path. This ensures that the proposed module can be applied independently to any robot mission, regardless of whether it is intended to find a radiation source. The method is validated in cross-condition randomized simulation environments and in real-world environments using an actual source and a robotic radiation detection system. The proposed method can serve as an advanced perception module, enhancing robotic radiation perception in complex environments.

\section*{Acknowledgments}
This material is based upon work supported by the Department of Energy, Office of Nuclear Energy, Distinguished Early
Career Program under Award Number DE-NE0009306.

\bibliographystyle{IEEEtran}
\bibliography{bibliography}

\end{document}